\documentclass{article} % For LaTeX2e
\usepackage{iclr2026_conference,times}

\usepackage{multirow}

\usepackage{hyperref}
\usepackage{url}
\usepackage{graphicx}
\newcommand{\methodName}{uncertainty distillation }
\usepackage{enumitem}
\usepackage{amsmath}
\usepackage{amsfonts}
\usepackage{times}
\usepackage{listings}

\colorlet{punct}{red!60!black}
\definecolor{background}{HTML}{EEEEEE}
\definecolor{delim}{RGB}{20,105,176}
\colorlet{numb}{magenta!60!black}

\lstdefinelanguage{json}{
    basicstyle=\normalfont\ttfamily,
    numbers=left,
    numberstyle=\scriptsize,
    stepnumber=1,
    numbersep=8pt,
    showstringspaces=false,
    breaklines=true,
    frame=lines,
    backgroundcolor=\color{background},
    literate=
     *{0}{{{\color{numb}0}}}{1}
      {1}{{{\color{numb}1}}}{1}
      {2}{{{\color{numb}2}}}{1}
      {3}{{{\color{numb}3}}}{1}
      {4}{{{\color{numb}4}}}{1}
      {5}{{{\color{numb}5}}}{1}
      {6}{{{\color{numb}6}}}{1}
      {7}{{{\color{numb}7}}}{1}
      {8}{{{\color{numb}8}}}{1}
      {9}{{{\color{numb}9}}}{1}
      {:}{{{\color{punct}{:}}}}{1}
      {,}{{{\color{punct}{,}}}}{1}
      {\{}{{{\color{delim}{\{}}}}{1}
      {\}}{{{\color{delim}{\}}}}}{1}
      {[}{{{\color{delim}{[}}}}{1}
      {]}{{{\color{delim}{]}}}}{1},
}

\usepackage{latexsym}
\usepackage{booktabs}
\usepackage{caption}  % added to control spacing after figures
\usepackage{algorithm}
\usepackage{xcolor}
\usepackage{algpseudocode}
\def\equationautorefname~#1\null{(#1)\null}
\def\itemautorefname~#1\null{(#1)\null}
\def\sectionautorefname~#1\null{\S#1\null}
\def\subsectionautorefname~#1\null{\S#1\null}
\def\subsubsectionautorefname~#1\null{\S#1\null}
\title{Uncertainty Distillation: Teaching Language Models to Express Semantic Confidence}

\author{Sophia Hager, David Mueller, Kevin Duh \& Nicholas Andrews \\
Department of Computer Science\\
Johns Hopkins University \\
\texttt{\{shager2,noa\}@jhu.edu} \\
 }

\iclrfinalcopy % Uncomment for camera-ready version, but NOT for submission.
\begin{document}

\maketitle

\begin{abstract}
As large language models (LLMs) are increasingly used for factual question-answering, it becomes more important for LLMs to have the capability to communicate the likelihood that their answer is correct. For these verbalized expressions of uncertainty to be meaningful, they should reflect the error rates at the expressed level of confidence. However, when prompted to express confidence, the error rates of current LLMs are inconsistent with their communicated confidences, highlighting the need for uncertainty quantification methods. Many prior methods calculate \textit{lexical} uncertainty, estimating a model's confidence in the specific string it generated. In some cases, however, it may be more useful to estimate \textit{semantic} uncertainty, or the model's confidence in the answer regardless of how it is verbalized. We propose a simple procedure, \textbf{uncertainty distillation}, to teach an LLM to verbalize calibrated semantic confidences. Using held-out data to map initial uncertainty estimates to meaningful probabilities, we create examples annotated with verbalized probabilities for supervised fine-tuning. We find that our method yields verbalized confidences that correlate well with observed error rates, even when compared to strong baselines, some of which are more than twenty times slower at inference time. Additionally, we demonstrate that our method can be applied to black-box models that allow API-based fine-tuning, resulting in estimates of uncertainty that are both more effective and more efficient than any of our baselines.
\end{abstract}

\section{Introduction}

%\Nick{Sophia: can you please include a little bit more related work here? (And/or the later related work section.) We don't need a dedicated section, but I just want to include a few more, e.g. touching on miscalibration of existing LLMs.}
Advances in LLM research have led to instruction-tuned generative models with impressive capabilities on many challenging tasks \citep{openai2024gpt4technicalreport,jiang2023mistral7b,dubey2024llama3herdmodels}. While the flexibility and quality of these models is appealing, they may still hallucinate or give incorrect answers \citep{rawte-etal-2023-troubling,Bai2024HallucinationOM}. However, language models do not readily provide an interpretable measure of a model's likelihood of correctness. LLMs tend to produce poorly-calibrated confidences when prompted to do so, and are often confidently incorrect~\citep{xiong2024can}. Furthermore, the elicited confidences may be impacted in unexpected ways by the choice of prompt~\citep{sclar2023quantifying}, such as the interpretation of ``very confident'' being dependent on the wording of the prompt.  

There are several other approaches as an alternative to prompting. Models' token-level probabilities can be used to provide information as a measure of \emph{lexical} uncertainty, which gives information about the likelihood of a generated string. This is often useful; however, the same fact can be expressed in any number of ways---``Berlin's the capital of Germany'' or ``The capital of Germany is Berlin!'' or ``Die Hauptstadt Deutschlands ist Berlin''---all capturing the same meaning~\citep{kuhn2023semantic}. \emph{Semantic} uncertainty is therefore challenging to capture, as token-level probabilities are influenced by the phrasing of an answer just as much as the semantics of the answer itself. This issue is particularly challenging for models employing large vocabularies such as multilingual language models, language models employing byte or character-level tokenization, or when using LLMs that are prone to producing extraneous outputs~\citep{xue2021mt5massivelymultilingualpretrained,wang2024mambabyte}.

%Another approach is to attempt to use the token-level predictive distribution to derive a measure of confidence, but this distribution expresses the \emph{lexical} uncertainty of the model rather than the \emph{semantic} uncertainty. The same fact can be expressed in any number of ways---``Berlin's the capital of Germany'' or ``The capital of Germany is Berlin!'' or ``Die Hauptstadt Deutschlands ist Berlin''---all capturing the same meaning~\cite{kuhn2023semantic}. This issue is exacerbated in models employing large vocabularies such as multilingual language models, in language models employing byte or character-level tokenization, or when using LLMs that are prone to producing extraneous outputs~\cite{xue2021mt5massivelymultilingualpretrained,wang2024mambabyte}. %Furthermore, the predictive distribution of the language model is not directly interpretable, being the product of a sequence of small probabilities. 
% requiring a post-hoc transformation to a 
%LLMs can relay well-calibrated uncertainty if specifically fine-tuned to do so~\citep{lin2022teaching}. 
%However, while instruction-tuned LLMs seem to have better capability for self-knowledge\cite{yin-etal-2023-large}, 
% \begin{figure}[t]
% \centering
% \includegraphics[scale=.3]{figs/placeholder_fig_1.png}
%   \caption{Placeholder. I will keep working on coming up with some idea of what to put here.}\label{fig:intro}
% \end{figure}

\begin{figure*}[t]
\centering
\includegraphics[scale=.31]{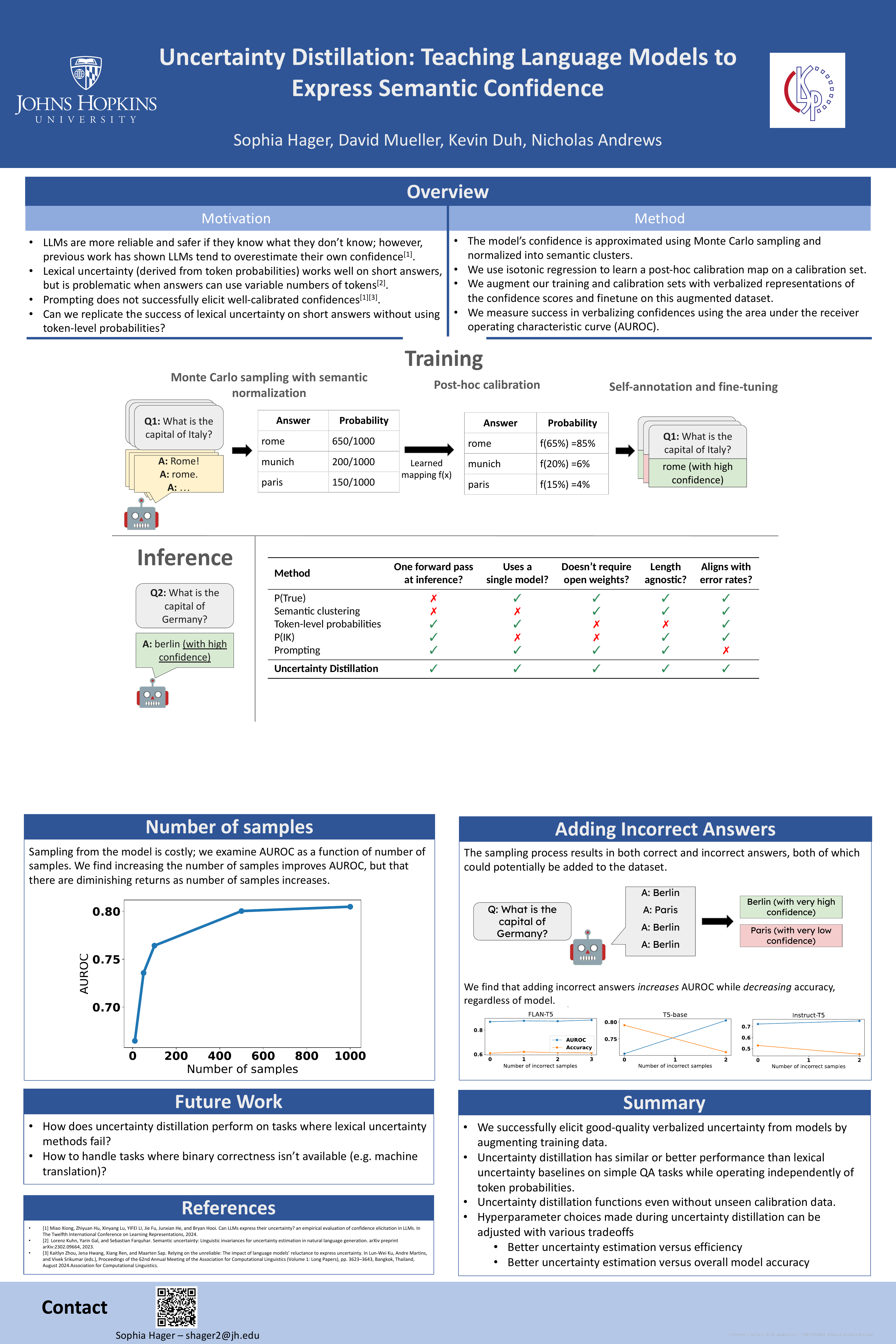}
    \captionsetup{belowskip=-10pt} % Set space below this caption
  \caption{An overview of our method, Uncertainty Distillation. At training time, in \textbf{Monte Carlo sampling with semantic normalization}, we sample repeatedly from our language model, and use a normalization function to consolidate answers with the same semantic meaning. By consolidating the counts, we obtain a Monte Carlo estimate of each answer's probability. In \textbf{post-hoc calibration}, we pass this estimate through a learned post-hoc calibration function to better align it with its likelihood of correctness. Finally, in \textbf{self-annotation and fine-tuning}, we translate these probabilities to verbalized signifiers and fine-tune a model to output verbalized confidences in addition to the answer. This method confers several advantages, listed in the table: at inference time, a single model generates the confidence efficiently in a single pass, providing high discriminative power with little computational overhead. The length of the answer does not directly impact the confidence, and white-box access to weights is not required.}\label{fig:method}
\end{figure*}
We present \emph{uncertainty distillation}\footnote{\textcolor{black}{We choose this name to evoke \emph{model} distillation, a process which like uncertainty distillation requires an offline cost to generate data to train a more efficient model.}}, a scheme for fine-tuning a language model to verbalize uncertainty based on its own internal state.
%\footnote{Our code and models are available at \url{https://github.com/<redacted>}}
Notably, \methodName teaches models to estimate their semantic---rather than lexical---uncertainty, as the distilled confidences are estimated from the probabilities of semantically normalized outputs, rather than relying on token-level probabilities. At inference time, models trained using \methodName efficiently generate a well-calibrated and interpretable statement of confidence in their answers, such as ``Berlin is the capital of Germany [high confidence].''\footnote{The uncertainty could be expressed in a variety of ways, including using special characters or numeric values.}
Our approach enables semantically equivalent but lexically different predictions to be assigned the same confidence, and a single generation with multiple claims can each be assigned different confidences.  Uncertainty distillation is computationally inexpensive at inference time, generating only a handful of additional tokens. Compared to methods such as \texttt{P(IK)}~\citep{Farquhar2024}, we do not require a separate uncertainty network; our approach uses standard supervised fine-tuning recipes for LLMs. Our method can be applied to open-source LLMs as well as proprietary LLMs that allow fine-tuning; \textcolor{black}{white-box} access to model weights is not required.

%Can we devise a training scheme that allows LLMs to produce well-calibrated estimates of uncertainty? %Constructing a static dataset to tune models to produce confidences presents several major difficulties. 
%In this work, we are interested in testable claims which can be compared to ground truth for correctness. However, such statements pose a challenge for uncertainty estimation with LLMs, which themselves are distributions over arbitrary strings. In particular, semantically equivalent statements could be expressed using different tokens in a generated answer, which introduces some complications in using token-level probabilities directly. 
%Thus, a first question to address is how to elicit a confidence score from an LLM, a choice we discuss further in~\autoref{sec:uncertainty}. Another major issue is that a model's uncertainty is particular to its own architecture and training; for instance, it would be unrealistic to assume that uncertainty scores extracted from a small LLaMa model\cite{dubey2024llama3herdmodels} would be  transferrable to GPT-4\cite{openai2024gpt4technicalreport}, or vice versa.
%Rather than constructing 
%In the typical instruction-tuning paradigm, one would produce supervised examples of input and outputs strings demonstrating the desired behavior.
%However, constructing a static dataset to tune models to produce confidences presents several major difficulties, the first being that the confidences must be faithful to the underlying LLM.
%Thus, we propose a scheme we call 

\emph{Uncertainty distillation} involves self-annotation of any desired QA dataset with the base model's calibrated uncertainties, which are then used to fine-tune that model to produce verbalized confidences. 
%To create training data, our approach samples from the model rather than using model probabilities, obviate potential issues revolving around tokenization schemes. 
At a high level (\autoref{fig:method}), our approach consists of three steps: (1) obtaining semantic uncertainty estimates from the model; (2) post-hoc calibrating these into meaningful probabilities; and (3) teaching the model via supervised fine-tuning to output verbalized confidences along with its predictions. 
%We are especially interested in understanding the impact of different ways of encoding verbalized confidences, and the extent to which verbalized confidences remain calibrated on held-out data.
%The calibrated posterior is then used to sample

%Diagnosing and correcting miscalibration requires held-out data where predicted events can be compared to observed events. In this work, we are interested in testable claims such as factual statements, which can be compared to ground truth for correctness. However, such statements pose a challenge for uncertainty estimation with LLMs, which themselves are distributions over arbitrary strings. In particular, semantically equivalent statements could be expressed using different tokens in a generated answer, which introduces some complications in using token-level probabilities directly. Thus, a first question to address is how to elicit a confidence score from an LLM, a choice we discuss further in~\autoref{sec:uncertainty}.

\vspace{-10pt}
\paragraph{Summary of contributions} 
\setlist{nolistsep}
    \begin{itemize}[noitemsep,leftmargin=*]
%\begin{itemize}
\item We propose uncertainty distillation, a simple yet effective scheme which uses supervised fine-tuning to teach LLMs to output calibrated semantic confidence statements along with their predictions. \textcolor{black}{We publish our code and trained models}.\footnote{\textcolor{black}{\url{https://anonymous.4open.science/r/uncertainty-distillation-anon-05CB/README.md} contains necessary code to replicate the results for API tuning.}}
\item We demonstrate that uncertainty distillation achieves easily interpretable results and compares favorably to several powerful baselines.
\item We analyze whether models trained with uncertainty distillation can apply their representations of uncertainty to unseen topics at inference time without further fine-tuning.
\end{itemize}
\section{Related Work}
\paragraph{Linguistic calibration and verbalized confidences} Generally, calibration refers to the concept that predicted probabilities should align with the probability of correctness \citep{Guo-etal-calibration-2017}. \citet{mielke-etal-2022-reducing} additionally propose the conception of ``linguistic calibration"---that models demonstrate uncertainty or doubt through natural language when they are incorrect, determining this uncertainty by using a predictor to determine the likelihood that an answer is correct and considering that to be the               model's uncertainty. There are significant advantages to verbalizing uncertainty: for one, there is relatively low computational overhead to generate several extra tokens, while using a separate calibration model to estimate confidence and then communicate this information to the user requires more computation at inference time~\citep{yang2024verbalizedconfidencescoresllms}. Verbalized confidences are also readily interpretable to an LLM when reasoning about uncertainty, or to an average end-user regardless of experience or background. 

\paragraph{Lexical uncertainty quantification}
Lexical uncertainty quantification metrics using information from token-level probabilities are commonly used and frequently effective \citep{hu2023uncertaintynaturallanguageprocessing, malinin2021uncertaintyestimationautoregressivestructured}. These probabilities are easily obtainable, do not require additional inference-time compute to generate, and often provide sufficient information for downstream use cases: e.g. error correction in chain of thought \citep{yin-etal-2024-reasoning}, hallucination detection \citep{arteaga2024hallucination}, or out-of-distribution data detection \citep{hendrycks2020pretrainedtransformersimproveoutofdistribution}. However, there are several disadvantages to lexical uncertainty quantification: it relies on model probabilities which may not be well-calibrated \citep{Guo-etal-calibration-2017}, and is often ineffective on calculating uncertainty of long generations \citep{zhang2024luqlongtextuncertaintyquantification}. The latter, in particular, may present problems for end users, as models trained using Reinforcement Learning from Human Feedback (RLHF) are often incentivized to produce long outputs \citep{singhal2024a}. It is therefore important to consider uncertainty quantification methods that do not rely on token-level probabilities to estimate uncertainty.

\paragraph{Semantic uncertainty quantification} 

In contexts where lexical uncertainty falls short, a natural method to obtain verbalized confidences might be to simply prompt a model to output confidences, providing an estimate of uncertainty without explicitly using token-level probabilities. However, in practice, LLMs tend to overestimate their own confidence, possibly because human annotators tend to prefer texts with fewer markers of uncertainty \citep{zhou-etal-2024-relying}. This, in turn, suggests while simply altering prompts may result in improved confidence estimates \citep{xiong2024can, tian-etal-2023-just}, models may be fundamentally limited in their ability to acknowledge uncertainty without further training.

Running multiple steps at inference time may provide a better estimate of semantic probability. \citet{xiong2024can} investigate several inference-time strategies which use multiple steps to estimate model uncertainty, such as sampling several answers on the same question or noting if a model changes its answer when prompted with a misleading alternative. While these methods do lead to improvements in LLM calibration, no single intervention consistently emerges as the most successful, and the authors note there is significant scope for improvement. \citet{kuhn2023semantic} and \citet{Farquhar2024} more explicitly relate this to semantic uncertainty, and find that sampling $m$ predictions from the model and clustering by semantic equivalence results in a robust measure of semantic uncertainty that compares favorably to lexical uncertainty. A major disadvantage of these sampling-based approaches is their increased computational complexity at inference time, however; for instance, the semantic clustering approach of~\citet{Farquhar2024}, which we compare to in our experiments, requires 20 samples and calls to a separate entailment model at inference time.
 
%In either case, it is necessary to estimate an LLM's uncertainty in its own predictions;  calibrating a model will only be as effective as the estimate of LLM uncertainty.

\section{Method}\label{sec:method}

We propose a simple training recipe, illustrated in \autoref{fig:method} and described below, to allow a language model to express confidences that correlate with expected error rates on held-out data. 

\subsection{Monte Carlo sampling with semantic normalization}\label{sec:uncertainty}

Assuming input $x$ and output $y$, we are looking to find $\sum_{y \in Y_{\text{equivalent}}}P(y \mid x)$, the model's likelihood of producing this answer or one that is semantically equivalent; however this would require marginalization over an infinite set of strings $Y$. To make this a tractable problem, we use a Monte Carlo approximation, where
%By sampling $k$ answers from the model, we estimate $P(y|x) \approx \frac{1}{k} \sum_{i=1}^{k} \{y_i = y\}$.  
 our estimate of the models' predictive distribution improves with $N$, at the expense of additional offline computation. Note however that we do not assume this quantity is a meaningful probability out-of-the-box due to potential overfitting or underfitting of the base model. To diagnose potential miscalibration of the base model as well as correct for it, we  \textcolor{black}{may fit a post-hoc calibrator if the training data demonstrates miscalibration.}

In more detail, to fit a post-hoc calibrator, we need a supervised dataset of datapoints not seen at training time $\{X^{\text{cal}}, Y^{\text{cal}}\}$. For each example $x \in X^{\text{cal}}$ we sample $N$ candidate answers $\{ \hat{y}_i\}_{i=1}^N \sim P_{\theta}(Y \mid X=x)$ from a model's predictive distribution\footnote{This model may have been fine-tuned on the specific task as in \autoref{sec:squad} or instruction-tuned as in \autoref{sec:experiments} and\autoref{sec:inst-tuning}.}. Before calculating the relative frequency of strings, we apply a normalization function (or set of normalization functions) to consolidate semantically similar outputs. In the short-form QA tasks we consider in~\autoref{sec:experiments}, we use the simple normalization function of isolating a multiple choice answer using tags, removing punctuation and standardizing capitalization; \textcolor{black}{we demonstrate how semantic normalization can be applied to more complex tasks in \autoref{sec:open-experiments}.} After consolidating strings belonging to the same event, the relative frequency $f$ of these events is a measure of the LLM's uncertainty in those events, although this may not be a well-calibrated probability.

\subsection{Post-hoc calibration}\label{sec:calibration}
   Neural networks are prone to miscalibration. A common remedy is to apply \emph{post-hoc} calibration methods, which usually involve some form of regression on predicted scores to transform them into meaningful probabilities. Specifically, we post-hoc calibrate the relative frequencies of each semantic cluster found in the previous step. Two common options for post-hoc calibration are isotonic regression and Platt scaling (sometimes called temperature scaling)~\citep{Guo-etal-calibration-2017}.  Our approach uses a model's predictions on $\{X^{\text{cal}}, Y^{\text{cal}}\}$ to diagnose and mitigate badly-calibrated initial model probabilities. We  fit an isotonic regression model\footnote{\textcolor{black}{We use isotonic regression for ease of training and use; this could be replaced with a different post-hoc calibration method, or omitted entirely as discussed in \autoref{sec:post-hoc}. We use the \texttt{scikit-learn 1.5.2} with no modification.}} on our calibration set by comparing the predicted scores to observed labels.\footnote{We discuss \textcolor{black}{the effect of} post-hoc calibration further in \autoref{sec:post-hoc}.} We compare each prediction $\hat{y}$ with score $f$ to observed events $y$. This yields a calibration map $c : \mathbb{R} \rightarrow [0, 1]$ we apply to the relative frequencies of events from samples in the previous step to yield probabilities.
\subsection{Self-annotation and fine-tuning}\label{sec:selfannote}
We compute the calibrated probability $p = c(f)$ associated with each prediction in the held-out calibration data, and choose a mapping into discrete confidence bins. Several options are possible for this binning function $b$, including adaptive schemes as well as uniform schemes, the number of bins $B$, and so on. In our experiments, we focus on a simple fixed-width scheme with $5$ bins.
Let $\hat{Y}$ denote the set of all predictions on $X^{\text{cal}}$, and, if the model was previously fine-tuned on a supervised training set $X^{\text{train}}$, we include predictions on $X^{\text{train}}$. We transform each prediction and calibrated confidence into a training example for a round of supervised fine-tuning by verbalizing the corresponding bin in the answer. For example, the fifth of five bins may correspond to ``very high confidence.'' The token sequences chosen to encode each bin are arbitrary, \textcolor{black}{as we discuss in \autoref{sec:binning};} for easy interpretability, we use short confidence descriptors in this paper, namely ``very low,'' ``low,'' ``medium,'' ``high,'' and ``very high.'' 

 In our scheme, we simply append the verbalized confidence to all answers. For instance, if the model generates 900 correct answers and 100 incorrect answers, there are two available data points that could potentially be added to the dataset:

\texttt{<correct answer> (with very high confidence)}

\texttt{<incorrect answer> (with very low confidence)}

While correct answers should be added as training data, appending the confidence scores to \textit{incorrect} answers may improve the model's ability to correctly verbalize its own confidence. However, it may also decrease the accuracy of the QA model. We introduce a hyperparameter to control the number of incorrect answers added to the training data. In~\autoref{sec:incorrect}, we further investigate the impact of this hyperparameter.

Starting from the sampled model, we perform supervised fine-tuning on these self-annotated targets with verbalized confidences to estimate a second model capable of verbalizing its confidence. If training an instruction-tuned model, we append an additional instruction such as ``Additionally state how confident you are in your answer.'' to the preexisting instruction\footnote{See \autoref{sec:prompts} for details on the specific prompts used in each experiment.}. If a reasoning trace has been generated during sampling, we randomly select a reasoning trace to add to the target answer from all possible options. At inference time, we obtain predictions \emph{and verbalized confidences} from this new model on held-out test data. \textcolor{black}{This test data has no overlap with the post-hoc calibrated training set, and can even be drawn from an entirely different dataset, as in \autoref{sec:analysis}.}  We remark that our model incurs little additional cost at inference time, as opposed to other confidence elicitation methods which require inference-time sampling~\citep{Farquhar2024,xiong2024can}.

\section{Experimental Setup}\label{sec:experiments}

\begin{table*}[h]
    \centering
    \begin{small}
    \begin{sc}
    \begin{tabular}{c|c|c|cc|cc}
    \toprule
          Dataset& Model & Method &AUROC& Acc& High Acc &High \%    \\
         \midrule
          \multirow{12}{*}{MMLU}&\multirow{6}{*}{Ministral-8B} &UD (ours) &\textbf{0.693}&0.601 & 0.766 &49.7\\
          &&Lexical baseline&0.627& 0.551& 0.555 & 99.2 \\
          &&Prompting&0.587&\textbf{0.637} & 0.643 &97.4 \\
          &&P(IK) & 0.670 &0.566&0.639&83.1\\
          &&P(True)&0.471&0.585&0.583&96.6\\
          &&Sem. Entropy &0.667 & 0.577&\textbf{0.821}&34.6\\
          \cmidrule{2-7}
          &\multirow{6}{*}{Llama-3B} &UD (ours) &\textbf{0.743}&0.532 & \textbf{0.759} &42.4 \\
          &&Lexical baseline&0.644&0.511&0.600&62.0\\
          &&Prompting&0.548&\textbf{0.613}&0.647&73.9\\
          &&P(IK)& 0.692&0.567&0.688&59.8\\
          &&P(True)&0.550&0.554&0.558&98.6\\
          &&Sem. Entropy & 0.646&0.560 &0.727&63.8\\
         \bottomrule 
         \multirow{12}{*}{SocialIQA}&\multirow{6}{*}{Ministral-8B} &UD (ours)&\textcolor{black}{0.671}&\textcolor{black}{0.713} & \textcolor{black}{\textbf{0.792}}&\textcolor{black}{53.7}\\
         &&Lexical baseline& 0.600& \textbf{0.738}& 0.760 &85.7\\
          &&Prompting&0.539&0.721 & 0.738 & 95.8\\
          &&P(IK)&\textbf{0.676}&0.650&0.713&85.0\\
          &&P(True)&0.491&0.712&0.710&92.5\\
          &&Sem. Entropy &0.603 &0.659 &0.780 &17.7\\
          \cmidrule{2-7}
          &\multirow{6}{*}{Llama-3B} &UD (ours)&\textbf{0.784}& 0.653& 0.833&55.1\\
          &&Lexical baseline& 0.531& 0.673& 0.687&95.3\\
          &&Prompting&0.545 & \textbf{0.685}& 0.712 &67.2 \\
          &&P(IK)&0.669&0.664&\textbf{0.839}&26.4\\
          &&P(True)&0.505&0.681&0.682&99.1\\
          &&Sem. Entropy & 0.601 & 0.675&0.758&34.0\\
         \bottomrule

    \end{tabular}
    \end{sc}
    \end{small}
    \caption{Binned AUROC and accuracy metrics for our large models and datasets. We find that uncertainty distillation (UD) leads to increased AUROC and accuracy in high-confidence categories. \texttt{Accuracy} is the overall accuracy, and \texttt{High Accuracy} is the accuracy for the most confident predictions. We find that uncertainty distillation with one generation achieves similar or improved \texttt{High Accuracy} compared to other methods, including those using multiple samples.}
    \label{tab:main}
\end{table*}

We examine the efficacy of uncertainty distillation in two settings. First, we demonstrate the success of uncertainty quantification with large language models trained on several standard QA benchmarks. Second, we examine whether the models can still accurately forecast uncertainty when applied to datasets not seen during uncertainty distillation.

\subsection{Uncertainty distillation in-domain}\label{sec:llm_UD}
\paragraph{Datasets} We demonstrate uncertainty distillation using two multiple-choice question answering datasets, the Massive Multitask Language Understanding benchmark (MMLU)~\citep{mmlu} and the Social Interaction Question Answering dataset (SocialIQA)~\citep{siqa}. MMLU consists of multiple choice questions over 57 subjects such as high school psychology or formal logic. We take a subset of 20,000 questions from the training set to act as our calibration data, a subset of 500 questions from the validation set to act as our validation data, and a subset of 2,000 quesions from the test set to act as our test data. SocialIQA is a dataset consisting of question/answer pairs about social situations. We take a subset of 20,000 questions from the training set to act as our calibration data, a subset of 500 questions from the training set to act as our validation data, and use the existing validation split as our test data. For both datasets we set $N=100$, i.e. we take 100 samples per question to construct our initial Monte Carlo estimate of confidence\footnote{\textcolor{black}{We chose $N$ based on the small-scale experiments described in \autoref{sec:samples}.}}

\paragraph{Models and baselines} We validate \methodName on these datasets using two modern instruction-tuned LLMs, Llama-3.2-3B-Instruct~\citep{dubey2024llama3herdmodels} and Ministral-8B-Instruct-2410~\citep{jiang2023mistral7b}. When performing uncertainty distillation with Ministral-8B, we use LoRA~\citep{hu-etal-lora-2021}.
 For the \texttt{Lexical} baseline, we extract token-level probabilities from the language model on our training/calibration split\footnote{As we do not have an initial fine-tuning step, these are equivalent.} and use this to train an isotonic regression model to calibrate the average token-level probability for each answer.\footnote{We use the average probability rather than the sequence probability to normalize over different lengths, as \citet{kuhn2023semantic} find this improves performance.} 
%We find that using this calibrator to map the average calibrated token probability of our test set to a confidence bin (e.g. 90\% to "high confidence") is successful in tasks with short answers, but fails with larger models which generate longer strings.
To measure the model's ability to verbalize its confidence prior to \methodName, in \texttt{Prompting} we prompt the base model to output its own confidence in its answer. We report this baseline for these models, and discuss the prompts used in \autoref{sec:prompts}. We also compare to \texttt{P(IK)} from \citet{Farquhar2024} which learns a mapping from hidden states to uncertainty scores, and \texttt{P(True)} from \citet{kadavath2022languagemodelsmostlyknow}. Finally, we compare to the \texttt{Semantic Entropy (SE)} approach from \citet{Farquhar2024}. Both \texttt{P(True)} and \texttt{Semantic Entropy} generate 20 samples from the model to compute uncertainty scores, unlike our approach which uses a single generation.
\subsection{Uncertainty distillation under domain shifts}
We have discussed uncertainty distillation as a method that allows a model to forecast its own certainty. However, one potential reason for its success is if it is instead learning information about the \textit{dataset}, and is learning to associate low confidence with types of questions that it has previously gotten wrong.\footnote{For instance, if models perform particularly poorly on chemistry questions, it might output low uncertainty only because the question uses words such as ``hydrogen'', rather than learning an innate representation of uncertainty.} By changing the evaluation dataset, we demonstrate that the representation of uncertainty is not limited to only the domain of the training dataset.

\paragraph{Datasets} We use SocialIQA and MMLU as described above. We also evaluate our models on the 500 examples in the test split of OpenbookQA\citep{openbook}, an elementary-level science multiple choice question answering dataset.

\paragraph{Models and baselines} In this experiments, we use the models described in \autoref{sec:llm_UD} without further fine-tuning. Models trained on MMLU are tested on SocialIQA and OpenbookQA; Models trained on SocialIQA are tested on MMLU and OpenbookQA. We compare to the \texttt{Lexical} and \texttt{P(IK)} baselines described above, as these are the only two methods that require supervised data (\texttt{Lexical} to fit a calibration map and \texttt{P(IK)} to train a regressor) and would be affected by domain shifts.

\subsection{Metrics} \label{sec:metrics}
We report the area under the receiver operating characteristic curve (AUROC),\footnote{Calculated using \texttt{scikit-learn 1.5.2}} which represents the probability that a randomly chosen correct answer will be in a higher-confidence bin than a randomly chosen incorrect answer. This metric is well established in previous literature (see e.g., \citet{hu2023uncertaintynaturallanguageprocessing}), and compares the relative rather than absolute probabilities, which allows us to use it effectively with discrete verbalized confidences.\footnote{We do not report  Expected Calibration Error (ECE), as it requires comparing a \textcolor{black}{numerical} probability to the prediction's true label, while our method and the semantic entropy baseline do not output \textcolor{black}{numerical probabilities}. Furthermore, many \textcolor{black}{forms of calibration error require the choice of several hyperparameters such as binning strategy or regularization, which can have a large impact on performance \citep{nixon2019measuring}.}}  Baseline methods that return a continuous score are binned to five categories to represent converting to a comparable verbalized confidence\footnote{\textcolor{black}{If the probability is not normalized, we learn a binner using the range from validation data.}}. For all methods, we plot the percentage of accurate answers in each bin to examine if confidence corresponds well with accuracy. We also report overall model accuracy, to evaluate the tradeoff between accuracy and calibration. Finally, we report \texttt{high accuracy} (accuracy of predictions in ``very high'' and ``high'' bins) and \texttt{high \%} (percentage of predictions in ``very high'' and ``high'' bins).  As an established use-case for verbalized confidences is to reject lower-confidence predictions, this provides information about how useful the LLM's predictions in rejecting incorrect answers and preserving a high number of correct answers.\footnote{The fact that high accuracy is not perfect also highlights a risk of confidence estimation: namely, that it increases trust in an answer that still may be incorrect.}

\begin{figure*}[h]
\centering
\includegraphics[scale=.21]{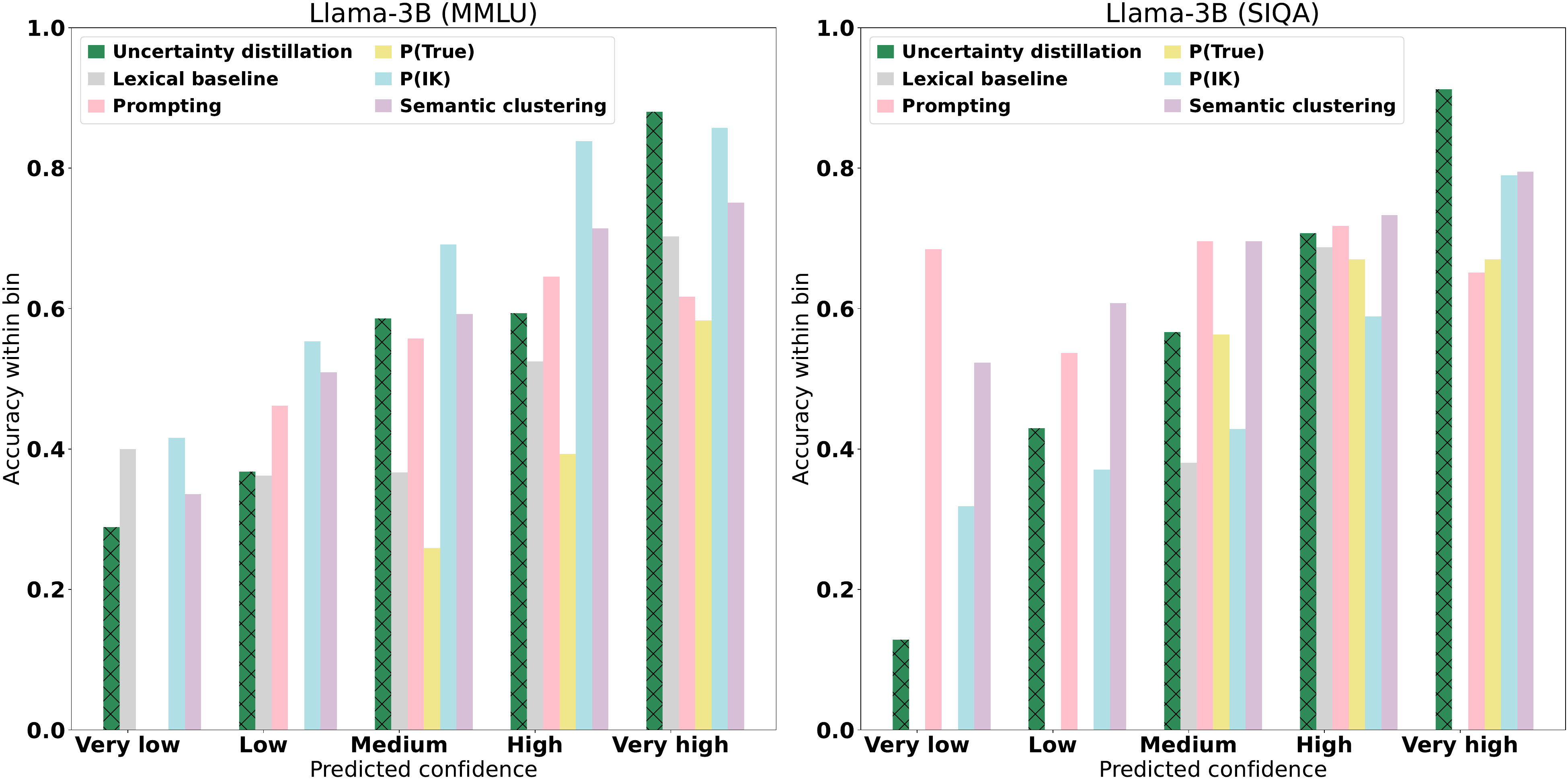}
  \caption{Average accuracy within each confidence bin for our experiments with Llama (Mistral results in \autoref{fig:mistral}). We find that our confidence bins correspond well with accuracy within the bin, while our baselines may not exhibit similar correspondence. We do not plot bins with fewer than 10 samples.}\label{fig:main}
\end{figure*}
\section{Results and discussion}\label{sec:discussion}

\autoref{fig:main} shows some of our results comparing \methodName to the lexical uncertainty baseline in terms of average accuracies in each confidence bin\footnote{We present the remaining two settings in \autoref{sec:more_plots}.}. In plots like this, an ideal model would exhibit a diagonal trend line where outputs reported to have high confidence indeed have high accuracy, and those in the low confidence bins have lower accuracy. We find that the verbalized confidences produced by \methodName are highly \textit{interpretable}, with high correspondence between accuracy of answers within a bin and that bins confidence. In contrast, confidence scores generated by the baselines may not correspond well with the actual accuracies within that bin. For instance, accuracy within the lowest confidence bin for the prompting baseline is 0.684 with Llama-3B on SocialIQA, while accuracy within the highest confidence bin is 0.651.

\autoref{tab:main} summarizes these plots in terms of AUROC score. AUROC is consistently high with uncertainty distillation, generally outperforming other methods.
We conclude that \methodName is effective for estimating confidence in an answer. AUROC is highest for \methodName for all experiments except Ministral-8B on SocialIQA, where it outperforms all baselines bu \texttt{P(IK)}. In particular, we note that uncertainty distillation consistently achieves higher AUROC than semantic entropy\citep{kuhn2023semantic}, despite semantic entropy requiring 20 samples and a computationally intensive clustering step at inference time: for instance, uncertainty distillation achieves AUROC of 0.784 with Llama-3B on SocialIQA, while semantic entropy achieves AUROC of 0.601.

The table also reports the accuracy of the highest confidence bin and the overall accuracy across all bins. While AUROC is the main metric for assessing performance, accuracy is also useful for understanding the nuances of the result. We find that uncertainty distillation does not lead to notable drops in overall accuracy, and that accuracy in the highest bins increases dramatically without restricting to drastically low amount high-confidence predictions (\texttt{High \%} stays consistently above 40\%). Uncertainty distillation achieves the best \texttt{High Accuracy} most cases.  The exceptions are Ministral-8B on MMLU and Llama-3B on SocialIQA. \textcolor{black}{In both these cases, the comparatively strong high accuracy results from notably smaller percentage of samples in high-confidence bins for these baselines, with only 34.6\% of predictions being high-confidence for the baseline for MMLU compared to 49.7\% for uncertainty distillation, and only 26.4\% of predictions being high-confidence for the baseline for SocialIQA, compared to 55.1\% for uncertainty distillation.}

\section{Success under domain shifts}\label{sec:analysis}

 \autoref{tab:gen} shows uncertainty distillation results compared to supervised baselines. We find that uncertainty distillation (UD) consistently achieves high AUROC despite the domain shifts, outperforming in all cases but Ministral-8B trained on SocialIQA and tested on OpenbookQA, which is outperformed by the lexical baseline and marginally by \texttt{P(IK)}. 

In \autoref{tab:gen}, we compare only to similarly out-of-domain baselines (i.e., also fit on data from a different distribution). A priori, one might expect that our approach fine-tuned for a specific dataset would significantly degrade in performance on a different dataset due to biases or spurious correlation. However, we find that out-of-domain uncertainty distillation outperforms all unsupervised baselines (semantic entropy, prompting, and \texttt{P(True))}, with the sole exception of Ministral-8B semantic entropy on MMLU. Notably, semantic entropy requires 20 samples from the language model, making uncertainty distillation more efficient at inference time by an order of magnitude. This result demonstrates that the representations of uncertainty learned by the model during uncertainty distillation are not limited to the training dataset, but can be applied to new datasets while still outperforming baselines unaffected by domain shifts. 
\begin{table*}[h]
    \centering
    \begin{small}
    \begin{sc}
    \begin{tabular}{c|c|c|c|cc}
    \toprule
          Train Dataset& Test Dataset & Model &Method&AUROC& Acc  \\
         \midrule
          \multirow{12}{*}{MMLU}&\multirow{6}{*}{SocialIQA} &\multirow{3}{*}{Ministral-8B}&UD (ours)&\textbf{0.657}&0.676 \\
          &&&Lexical baseline&0.593&\textbf{0.738} \\
          &&&P(IK)&0.618&0.636\\
          \cmidrule{3-6}

          &&\multirow{3}{*}{Llama-3B} &UD (ours)& \textbf{0.717}&0.627\\
          &&&Lexical baseline &0.574&\textbf{0.670}\\
          &&&P(IK) &0.675&0.655\\
          \cmidrule{2-6}
          &\multirow{6}{*}{OpenbookQA} &\multirow{3}{*}{Ministral-8B}&UD (ours) &\textbf{0.757}&0.734\\
          &&&Lexical baseline&0.676&\textbf{0.812} \\
          &&&P(IK)&0.683&0.736\\

          \cmidrule{3-6}
          &&\multirow{3}{*}{Llama-3B} &UD (ours) &\textbf{0.834}&\textbf{0.733} \\
          &&&Lexical baseline &0.647&0.680\\
          &&&P(IK)&0.770&0.722 \\
          \midrule
          \midrule

          \multirow{12}{*}
          {SocialIQA}&\multirow{6}{*}{MMLU} &\multirow{3}{*}{Ministral-8B}&UD (ours)&\textbf{\textcolor{black}{0.644}}&\textbf{\textcolor{black}{0.599}} \\
          &&&Lexical baseline& 0.635&0.551 \\
          &&&P(IK)&0.605&0.553\\
          \cmidrule{3-6}

          &&\multirow{3}{*}{Llama-3B} &UD (ours)& \textbf{0.714}& 0.547\\
          &&&Lexical baseline  &0.569&0.528\\
          &&&P(IK)&0.687&\textbf{0.572} \\
          \cmidrule{2-6}
          &\multirow{6}{*}{OpenbookQA} &\multirow{3}{*}{Ministral-8B}&UD (ours)&\textcolor{black}{0.700} &\textcolor{black}{0.746} \\
          &&&Lexical baseline&\textbf{0.719}&\textbf{0.812}\\
          &&&P(IK)&0.704&0.718\\
          \cmidrule{3-6}
          &&\multirow{3}{*}{Llama-3B} &UD (ours) & \textbf{0.758} & \textbf{0.755}\\
          
          &&&Lexical baseline &0.549 &0.680\\
          &&&P(IK) &0.693&0.694 \\
          \midrule

    \end{tabular}
    \end{sc}
    \end{small}
    \caption{AUROC and accuracy metrics for Uncertainty Distillation (UD) tested on  out-of-domain datasets compared to out-of-domain supervised baselines tested. Uncertainty distillation consistently achieve high AUROC on the novel test set in comparison to the supervised baselines, which are more inconsistent when dealing with domain shifts.}
    \label{tab:gen}
\end{table*}
\section{\textcolor{black}{Uncertainty distillation with black-box models} }\label{sec:api-tuning}
\color{black}
Increasingly, large foundation models are not being released publicly, and even if they were, few groups posses the hardware to run large mixture-of-experts models efficiently.  One advantage of uncertainty distillation is that it does not require open access to model weights; therefore, if there is an option to tune a model through an API, uncertainty distillation can still be used. Here, we demonstrate the success of uncertainty distillation in this case. %In this setting to prioritize cost-effectiveness, we use only DETAILS samples to generate our training data.

\paragraph{Model and dataset} To strike a balance between cost and quality, we use Google's \texttt{gemini-2.5-flash-lite} model. Since this is a significantly more capable model than the open-weight models used elsewhere in this paper, we use a more challenging benchmark: MMLU-Pro~\citep{wang2024mmluprorobustchallengingmultitask}, a variant of MMLU designed to be more challenging and which includes a broader set of questions, including questions requiring reasoning. Note that the original MMLU dataset already covers a wide range of topics, so this benchmark helps us understand whether a single model can successfully estimate uncertainty across a wide range of settings, given only a a few hundred demonstrations from each domain.  We use an existing split of the data into training and evaluation sets, and we further split the evaluation set into 50\% validation data and 50\% test data\footnote{\url{https://huggingface.co/datasets/answerdotai/MMLU-SemiPro}}.

\paragraph{Baselines} The API restrictions preclude uncertainty estimation approaches that inspect model activations such as \texttt{P(IK)} or approaches that require next-token logits such as the lexical baseline. Nonetheless, we can fairly compare to baselines involving prompting or repeated sampling, so we include comparisons to prompting for verbalized confidences and semantic entropy. For semantic entropy, we include results for $8$, $16$, and $32$ samples at inference time.

\paragraph{Procedure} The procedure is identical using a commercial API or fine-tuning models locally. First, we generate $128$ samples on the training split and then apply semantic clustering to estimate the relative frequency of each prediction. We then post-hoc calibrate the relative frequencies using either temperature scaling or isotonic regression. Finally, we create a fine-tuning dataset consisting of predictions and their calibrated confidences. On the validation data, we compare the performance of models trained with varying numbers of incorrect predictions, as described in~\autoref{sec:selfannote} and~\autoref{sec:hparams}. The base model is then fine-tuned using LoRA via the Google Generative AI SDK\footnote{\url{https://docs.cloud.google.com/vertex-ai/generative-ai/docs/models/gemini-use-supervised-tuning}}, and this fine-tuned model is then used to make predictions on validation or test data. 

\paragraph{Results and analysis}
The cost of running the entire pipeline was approximately \$20, including generating samples, fine-tuning, and generating predictions on held-out data. For experimenting with different fine-tuning hyper-parameters, we could re-use  samples, further controlling costs.  We show results in \autoref{tab:api}, demonstrating that uncertainty distillation outperforms the other black-box methods. In particular, we note that at inference uncertainty distillation only requires a single pass, while semantic entropy requires eight to thirty-two. 
\begin{table*}[h]
    \centering
    \begin{small}
    \begin{sc}
    \begin{tabular}{c|cc|c|c}
    \toprule
          \color{black}Method &\color{black}AUROC& \color{black}Acc&\color{black} High Acc  & \color{black}Cost per answer   \\
         \midrule
          \color{black}UD (ours) &\color{black}\textbf{0.762}&\color{black}0.490&\color{black} \textbf{0.706} & \color{black}1x \\
          
          \color{black}Prompting&\color{black}0.582&\color{black}0.498&\color{black} 0.503 & \color{black} \color{black}1x \\
          \color{black}Sem. Entropy (8)&\color{black}0.713 &\color{black}\textbf{0.508} &\color{black}0.562 & \color{black}8x \\
          \color{black}Sem. Entropy (16)&\color{black}0.715 &\color{black}0.505 &\color{black}0.575 & \color{black}16x\\
          \color{black}Sem. Entropy (32)&\color{black}0.718 &\color{black}0.505 &\color{black}0.581 & \color{black}32x\\
          \bottomrule

    \end{tabular}
    \end{sc}
    \end{small}
    \caption{\color{black}AUROC and accuracy metrics for the API-tuning experiments for \texttt{gemini-2.5-flash-lite}. We find that uncertainty distillation (UD) significantly outperforms all baselines in AUROC and high accuracy, and achieves similar accuracy to the only single-generation baseline, prompting. With the multi-generation baseline of semantic entropy increasing the number of samples to 32 does not cause semantic entropy to approach the performance of uncertainty distillation. We also note that semantic entropy costs 8-32x more than uncertainty distillation.}
    \label{tab:api}
\end{table*}

The limitations in applicable baselines demonstrate an appealing feature of uncertainty distillation; specifically, that for black-box models such as Gemini it is possible to achieve high performance better than semantic entropy with the efficiency of prompting, while most other accurate uncertainty quantification measures cannot be applied without open access to model weights.
\color{black}

\section{Conclusion}\label{sec:conclusion}

\paragraph{Findings} We find that \methodName leads to improved estimates of uncertainty in comparison to many strong baselines, including baselines that require considerably more samples at inference-time. Additionally, we demonstrate that the representations of uncertainty learned during uncertainty distillation are applicable to unfamiliar test sets, showing that the model is learning to predict its own uncertainty independent of the subject of the dataset. Overall, we view our contribution as a significant step towards LLMs that can reliably reason about uncertainty, without requiring any auxiliary models or incurring additional inference-time compute.

\paragraph{Future work} %We focus on strings that are quickly and easily normalized with preprocessing functions, but 
While we focus on QA tasks, our method could be applied to tasks outside simple QA through the use of LLM verifiers to calculate binary correctness, as discussed in \autoref{sec:semantic}. Future work may also investigate the robustness of the model's internal representation of uncertainty to even more dramatic domain shifts, such as different types of QA tasks or even tasks such as machine translation that bear no similarity to question answering. Looking beyond these immediate questions, LLMs that are able to verbalize meaningful confidences, for example thanks to our method, may be useful in a variety of applications requiring reasoning about uncertainty, such as medical diagnosis.

% A higher-level question is whether model size affects a model's ability to learn its own uncertainty. If larger models can learn to express what they do not know, it may be possible to investigate whether there is multi-task transfer of this ability; for instance, does improved uncertainty prediction in the domain of question answering lead to improvements in a related task such as question understanding?

\section*{Limitations}
%We run our method on relatively simple QA tasks, where computing uncertainty is comparatively straightforward. Our method is also not infallible, with imperfect accuracies in high-confidence bins; incorrect high confidence predictions may present a small risk of prompting users to trust predictions more than they otherwise would have.
Our experiments focus on established QA tasks which admit straightforward ways to assess correctness.
In principle, our approach generalizes to more complex tasks involving longer-form generations
\textcolor{black}{than the open-answer QA task described here; we leave it as future work to experiment in these settings with LLM verification. Separately, the proposed approach may be useful in cases where a single generation involves multiple distinct claims that each need to be associated with distinct confidences. Future work should identify appropriate datasets to evaluate multi-claim uncertainty estimation. We hope that our findings will encourage further study into \methodName in more general settings.}%, for example using an LLM verifier to establish correctness; we leave it as future work to experiment in these settings. Separately, the proposed approach may be useful in cases where a single generation involves multiple distinct claims that each need to be associated with distinct confidences. Future work should identify appropriate datasets to evaluate multi-claim uncertainty estimation. Finally, our experiments do not include models larger than 8 billion parameters due to compute limitations, and are performed entirely on open-source models rather than fine-tuning through proprietary APIs. However, we hope that our findings will encourage further study into \methodName for larger LLMs and in more general settings.

\section*{Reproducibility statement}
We have endeavored to make reproducing our results straightforward. We describe our datasets, models, and metrics in detail in \autoref{sec:llm_UD}; we provide the prompts used in \autoref{sec:prompts}; we provide the used hyperparameters in \autoref{sec:hparams} and \autoref{sec:instruct-hparams}; and we report the compute resources and dataset licensing in \autoref{sec:resources}. \textcolor{black}{We plan to release our code upon publication.}
%\section*{Acknowledgments}

\bibliography{iclr2026_conference}

@article{wang2024mambabyte,
  title={Mambabyte: Token-free selective state space model},
  author={Wang, Junxiong and Gangavarapu, Tushaar and Yan, Jing Nathan and Rush, Alexander M},
  journal={arXiv preprint arXiv:2401.13660},
  year={2024}
}

@misc{malinin2021uncertaintyestimationautoregressivestructured,
      title={Uncertainty Estimation in Autoregressive Structured Prediction}, 
      author={Andrey Malinin and Mark Gales},
      year={2021},
      eprint={2002.07650},
      archivePrefix={arXiv},
      primaryClass={stat.ML},
      url={https://arxiv.org/abs/2002.07650}, 
}

@inproceedings{yin-etal-2024-reasoning,
    title = "Reasoning in Flux: Enhancing Large Language Models Reasoning through Uncertainty-aware Adaptive Guidance",
    author = "Yin, Zhangyue  and
      Sun, Qiushi  and
      Guo, Qipeng  and
      Zeng, Zhiyuan  and
      Li, Xiaonan  and
      Dai, Junqi  and
      Cheng, Qinyuan  and
      Huang, Xuanjing  and
      Qiu, Xipeng",
    editor = "Ku, Lun-Wei  and
      Martins, Andre  and
      Srikumar, Vivek",
    booktitle = "Proceedings of the 62nd Annual Meeting of the Association for Computational Linguistics (Volume 1: Long Papers)",
    month = aug,
    year = "2024",
    address = "Bangkok, Thailand",
    publisher = "Association for Computational Linguistics",
    url = "https://aclanthology.org/2024.acl-long.131/",
    doi = "10.18653/v1/2024.acl-long.131",
    pages = "2401--2416"
}

@misc{zhang2024luqlongtextuncertaintyquantification,
      title={LUQ: Long-text Uncertainty Quantification for LLMs}, 
      author={Caiqi Zhang and Fangyu Liu and Marco Basaldella and Nigel Collier},
      year={2024},
      eprint={2403.20279},
      archivePrefix={arXiv},
      primaryClass={cs.CL},
      url={https://arxiv.org/abs/2403.20279}, 
}

@misc{yang2024verbalizedconfidencescoresllms,
      title={On Verbalized Confidence Scores for LLMs}, 
      author={Daniel Yang and Yao-Hung Hubert Tsai and Makoto Yamada},
      year={2024},
      eprint={2412.14737},
      archivePrefix={arXiv},
      primaryClass={cs.CL},
      url={https://arxiv.org/abs/2412.14737}, 
}

@inproceedings{
singhal2024a,
title={A Long Way to Go: Investigating Length Correlations in {RLHF}},
author={Prasann Singhal and Tanya Goyal and Jiacheng Xu and Greg Durrett},
booktitle={First Conference on Language Modeling},
year={2024},
url={https://openreview.net/forum?id=G8LaO1P0xv}
}

@misc{hendrycks2020pretrainedtransformersimproveoutofdistribution,
      title={Pretrained Transformers Improve Out-of-Distribution Robustness}, 
      author={Dan Hendrycks and Xiaoyuan Liu and Eric Wallace and Adam Dziedzic and Rishabh Krishnan and Dawn Song},
      year={2020},
      eprint={2004.06100},
      archivePrefix={arXiv},
      primaryClass={cs.CL},
      url={https://arxiv.org/abs/2004.06100}, 
}

@inproceedings{
arteaga2024hallucination,
title={Hallucination Detection in {LLM}s: Fast and Memory-Efficient Finetuned Models},
author={Gabriel Y. Arteaga and Thomas B. Sch{\"o}n and Nicolas Pielawski},
booktitle={Northern Lights Deep Learning Conference 2025},
year={2024},
url={https://openreview.net/forum?id=8T8QkDsuO9}
}

@misc{hu2023uncertaintynaturallanguageprocessing,
      title={Uncertainty in Natural Language Processing: Sources, Quantification, and Applications}, 
      author={Mengting Hu and Zhen Zhang and Shiwan Zhao and Minlie Huang and Bingzhe Wu},
      year={2023},
      eprint={2306.04459},
      archivePrefix={arXiv},
      primaryClass={cs.CL},
      url={https://arxiv.org/abs/2306.04459}, 
}

@misc{xue2021mt5massivelymultilingualpretrained,
      title={mT5: A massively multilingual pre-trained text-to-text transformer}, 
      author={Linting Xue and Noah Constant and Adam Roberts and Mihir Kale and Rami Al-Rfou and Aditya Siddhant and Aditya Barua and Colin Raffel},
      year={2021},
      eprint={2010.11934},
      archivePrefix={arXiv},
      primaryClass={cs.CL},
      url={https://arxiv.org/abs/2010.11934}, 
}

@article{DBLP:journals/corr/DuSC17,
  author       = {Xinya Du and
                  Junru Shao and
                  Claire Cardie},
  title        = {Learning to Ask: Neural Question Generation for Reading Comprehension},
  journal      = {CoRR},
  volume       = {abs/1705.00106},
  year         = {2017},
  url          = {http://arxiv.org/abs/1705.00106},
  eprinttype    = {arXiv},
  eprint       = {1705.00106},
  bibsource    = {dblp computer science bibliography, https://dblp.org}
}

@article{rajpurkar2016squad,
  title={Squad: 100,000+ questions for machine comprehension of text},
  author={Rajpurkar, P},
  journal={arXiv preprint arXiv:1606.05250},
  year={2016}
}

@article{kuhn2023semantic,
  title={Semantic uncertainty: Linguistic invariances for uncertainty estimation in natural language generation},
  author={Kuhn, Lorenz and Gal, Yarin and Farquhar, Sebastian},
  journal={arXiv preprint arXiv:2302.09664},
  year={2023}
}

@article{huang2024calibrating,
  title={Calibrating Long-form Generations from Large Language Models},
  author={Huang, Yukun and Liu, Yixin and Thirukovalluru, Raghuveer and Cohan, Arman and Dhingra, Bhuwan},
  journal={arXiv preprint arXiv:2402.06544},
  year={2024}
}

@article{sclar2023quantifying,
  title={Quantifying Language Models' Sensitivity to Spurious Features in Prompt Design or: How I learned to start worrying about prompt formatting},
  author={Sclar, Melanie and Choi, Yejin and Tsvetkov, Yulia and Suhr, Alane},
  journal={arXiv preprint arXiv:2310.11324},
  year={2023}
}

@article{raffel2020exploring,
  title={Exploring the limits of transfer learning with a unified text-to-text transformer},
  author={Raffel, Colin and Shazeer, Noam and Roberts, Adam and Lee, Katherine and Narang, Sharan and Matena, Michael and Zhou, Yanqi and Li, Wei and Liu, Peter J},
  journal={Journal of machine learning research},
  volume={21},
  number={140},
  pages={1--67},
  year={2020}
}

@inproceedings{nixon2019measuring,
  title={Measuring Calibration in Deep Learning.},
  author={Nixon, Jeremy and Dusenberry, Michael W and Zhang, Linchuan and Jerfel, Ghassen and Tran, Dustin},
  booktitle={CVPR workshops},
  volume={2},
  number={7},
  year={2019}
}

@inproceedings{zhou-etal-2024-relying,
    title = "Relying on the Unreliable: The Impact of Language Models{'} Reluctance to Express Uncertainty",
    author = "Zhou, Kaitlyn  and
      Hwang, Jena  and
      Ren, Xiang  and
      Sap, Maarten",
    editor = "Ku, Lun-Wei  and
      Martins, Andre  and
      Srikumar, Vivek",
    booktitle = "Proceedings of the 62nd Annual Meeting of the Association for Computational Linguistics (Volume 1: Long Papers)",
    month = aug,
    year = "2024",
    address = "Bangkok, Thailand",
    publisher = "Association for Computational Linguistics",
    url = "https://aclanthology.org/2024.acl-long.198",
    pages = "3623--3643",
}

@inproceedings{
xiong2024can,
title={Can {LLM}s Express Their Uncertainty? An Empirical Evaluation of Confidence Elicitation in {LLM}s},
author={Miao Xiong and Zhiyuan Hu and Xinyang Lu and YIFEI LI and Jie Fu and Junxian He and Bryan Hooi},
booktitle={The Twelfth International Conference on Learning Representations},
year={2024},
url={https://openreview.net/forum?id=gjeQKFxFpZ}
}

@article{mielke-etal-2022-reducing,
    title = "Reducing Conversational Agents{'} Overconfidence Through Linguistic Calibration",
    author = "Mielke, Sabrina J.  and
      Szlam, Arthur  and
      Dinan, Emily  and
      Boureau, Y-Lan",
    editor = "Roark, Brian  and
      Nenkova, Ani",
    journal = "Transactions of the Association for Computational Linguistics",
    volume = "10",
    year = "2022",
    address = "Cambridge, MA",
    publisher = "MIT Press",
    url = "https://aclanthology.org/2022.tacl-1.50",
    doi = "10.1162/tacl_a_00494",
    pages = "857--872",
}

@inproceedings{Guo-etal-calibration-2017,
author = {Guo, Chuan and Pleiss, Geoff and Sun, Yu and Weinberger, Kilian Q.},
title = {On calibration of modern neural networks},
year = {2017},
publisher = {JMLR.org},
booktitle = {Proceedings of the 34th International Conference on Machine Learning - Volume 70},
pages = {1321–1330},
numpages = {10},
location = {Sydney, NSW, Australia},
series = {ICML'17}
}

@inproceedings{tian-etal-2023-just,
    title = "Just Ask for Calibration: Strategies for Eliciting Calibrated Confidence Scores from Language Models Fine-Tuned with Human Feedback",
    author = "Tian, Katherine  and
      Mitchell, Eric  and
      Zhou, Allan  and
      Sharma, Archit  and
      Rafailov, Rafael  and
      Yao, Huaxiu  and
      Finn, Chelsea  and
      Manning, Christopher",
    editor = "Bouamor, Houda  and
      Pino, Juan  and
      Bali, Kalika",
    booktitle = "Proceedings of the 2023 Conference on Empirical Methods in Natural Language Processing",
    month = dec,
    year = "2023",
    address = "Singapore",
    publisher = "Association for Computational Linguistics",
    url = "https://aclanthology.org/2023.emnlp-main.330",
    doi = "10.18653/v1/2023.emnlp-main.330",
    pages = "5433--5442",
}

@article{openbook,
  author       = {Todor Mihaylov and
                  Peter Clark and
                  Tushar Khot and
                  Ashish Sabharwal},
  title        = {Can a Suit of Armor Conduct Electricity? {A} New Dataset for Open
                  Book Question Answering},
  journal      = {CoRR},
  volume       = {abs/1809.02789},
  year         = {2018},
  url          = {http://arxiv.org/abs/1809.02789},
  eprinttype    = {arXiv},
  eprint       = {1809.02789},
  bibsource    = {dblp computer science bibliography, https://dblp.org}
}

@misc{openai2024gpt4technicalreport,
      title={GPT-4 Technical Report}, 
      author={OpenAI and Josh Achiam and Steven Adler and Sandhini Agarwal and Lama Ahmad and Ilge Akkaya and Florencia Leoni Aleman and Diogo Almeida and Janko Altenschmidt and Sam Altman and Shyamal Anadkat and Red Avila and Igor Babuschkin and Suchir Balaji and Valerie Balcom and Paul Baltescu and Haiming Bao and Mohammad Bavarian and Jeff Belgum and Irwan Bello and Jake Berdine and Gabriel Bernadett-Shapiro and Christopher Berner and Lenny Bogdonoff and Oleg Boiko and Madelaine Boyd and Anna-Luisa Brakman and Greg Brockman and Tim Brooks and Miles Brundage and Kevin Button and Trevor Cai and Rosie Campbell and Andrew Cann and Brittany Carey and Chelsea Carlson and Rory Carmichael and Brooke Chan and Che Chang and Fotis Chantzis and Derek Chen and Sully Chen and Ruby Chen and Jason Chen and Mark Chen and Ben Chess and Chester Cho and Casey Chu and Hyung Won Chung and Dave Cummings and Jeremiah Currier and Yunxing Dai and Cory Decareaux and Thomas Degry and Noah Deutsch and Damien Deville and Arka Dhar and David Dohan and Steve Dowling and Sheila Dunning and Adrien Ecoffet and Atty Eleti and Tyna Eloundou and David Farhi and Liam Fedus and Niko Felix and Simón Posada Fishman and Juston Forte and Isabella Fulford and Leo Gao and Elie Georges and Christian Gibson and Vik Goel and Tarun Gogineni and Gabriel Goh and Rapha Gontijo-Lopes and Jonathan Gordon and Morgan Grafstein and Scott Gray and Ryan Greene and Joshua Gross and Shixiang Shane Gu and Yufei Guo and Chris Hallacy and Jesse Han and Jeff Harris and Yuchen He and Mike Heaton and Johannes Heidecke and Chris Hesse and Alan Hickey and Wade Hickey and Peter Hoeschele and Brandon Houghton and Kenny Hsu and Shengli Hu and Xin Hu and Joost Huizinga and Shantanu Jain and Shawn Jain and Joanne Jang and Angela Jiang and Roger Jiang and Haozhun Jin and Denny Jin and Shino Jomoto and Billie Jonn and Heewoo Jun and Tomer Kaftan and Łukasz Kaiser and Ali Kamali and Ingmar Kanitscheider and Nitish Shirish Keskar and Tabarak Khan and Logan Kilpatrick and Jong Wook Kim and Christina Kim and Yongjik Kim and Jan Hendrik Kirchner and Jamie Kiros and Matt Knight and Daniel Kokotajlo and Łukasz Kondraciuk and Andrew Kondrich and Aris Konstantinidis and Kyle Kosic and Gretchen Krueger and Vishal Kuo and Michael Lampe and Ikai Lan and Teddy Lee and Jan Leike and Jade Leung and Daniel Levy and Chak Ming Li and Rachel Lim and Molly Lin and Stephanie Lin and Mateusz Litwin and Theresa Lopez and Ryan Lowe and Patricia Lue and Anna Makanju and Kim Malfacini and Sam Manning and Todor Markov and Yaniv Markovski and Bianca Martin and Katie Mayer and Andrew Mayne and Bob McGrew and Scott Mayer McKinney and Christine McLeavey and Paul McMillan and Jake McNeil and David Medina and Aalok Mehta and Jacob Menick and Luke Metz and Andrey Mishchenko and Pamela Mishkin and Vinnie Monaco and Evan Morikawa and Daniel Mossing and Tong Mu and Mira Murati and Oleg Murk and David Mély and Ashvin Nair and Reiichiro Nakano and Rajeev Nayak and Arvind Neelakantan and Richard Ngo and Hyeonwoo Noh and Long Ouyang and Cullen O'Keefe and Jakub Pachocki and Alex Paino and Joe Palermo and Ashley Pantuliano and Giambattista Parascandolo and Joel Parish and Emy Parparita and Alex Passos and Mikhail Pavlov and Andrew Peng and Adam Perelman and Filipe de Avila Belbute Peres and Michael Petrov and Henrique Ponde de Oliveira Pinto and Michael and Pokorny and Michelle Pokrass and Vitchyr H. Pong and Tolly Powell and Alethea Power and Boris Power and Elizabeth Proehl and Raul Puri and Alec Radford and Jack Rae and Aditya Ramesh and Cameron Raymond and Francis Real and Kendra Rimbach and Carl Ross and Bob Rotsted and Henri Roussez and Nick Ryder and Mario Saltarelli and Ted Sanders and Shibani Santurkar and Girish Sastry and Heather Schmidt and David Schnurr and John Schulman and Daniel Selsam and Kyla Sheppard and Toki Sherbakov and Jessica Shieh and Sarah Shoker and Pranav Shyam and Szymon Sidor and Eric Sigler and Maddie Simens and Jordan Sitkin and Katarina Slama and Ian Sohl and Benjamin Sokolowsky and Yang Song and Natalie Staudacher and Felipe Petroski Such and Natalie Summers and Ilya Sutskever and Jie Tang and Nikolas Tezak and Madeleine B. Thompson and Phil Tillet and Amin Tootoonchian and Elizabeth Tseng and Preston Tuggle and Nick Turley and Jerry Tworek and Juan Felipe Cerón Uribe and Andrea Vallone and Arun Vijayvergiya and Chelsea Voss and Carroll Wainwright and Justin Jay Wang and Alvin Wang and Ben Wang and Jonathan Ward and Jason Wei and CJ Weinmann and Akila Welihinda and Peter Welinder and Jiayi Weng and Lilian Weng and Matt Wiethoff and Dave Willner and Clemens Winter and Samuel Wolrich and Hannah Wong and Lauren Workman and Sherwin Wu and Jeff Wu and Michael Wu and Kai Xiao and Tao Xu and Sarah Yoo and Kevin Yu and Qiming Yuan and Wojciech Zaremba and Rowan Zellers and Chong Zhang and Marvin Zhang and Shengjia Zhao and Tianhao Zheng and Juntang Zhuang and William Zhuk and Barret Zoph},
      year={2024},
      eprint={2303.08774},
      archivePrefix={arXiv},
      primaryClass={cs.CL},
      url={https://arxiv.org/abs/2303.08774}, 
}

@misc{jiang2023mistral7b,
      title={Mistral 7B}, 
      author={Albert Q. Jiang and Alexandre Sablayrolles and Arthur Mensch and Chris Bamford and Devendra Singh Chaplot and Diego de las Casas and Florian Bressand and Gianna Lengyel and Guillaume Lample and Lucile Saulnier and Lélio Renard Lavaud and Marie-Anne Lachaux and Pierre Stock and Teven Le Scao and Thibaut Lavril and Thomas Wang and Timothée Lacroix and William El Sayed},
      year={2023},
      eprint={2310.06825},
      archivePrefix={arXiv},
      primaryClass={cs.CL},
      url={https://arxiv.org/abs/2310.06825}, 
}

@misc{dubey2024llama3herdmodels,
      title={The Llama 3 Herd of Models}, 
      author={Abhimanyu Dubey and Abhinav Jauhri and Abhinav Pandey and Abhishek Kadian and Ahmad Al-Dahle and Aiesha Letman and Akhil Mathur and Alan Schelten and Amy Yang and Angela Fan and Anirudh Goyal and Anthony Hartshorn and Aobo Yang and Archi Mitra and Archie Sravankumar and Artem Korenev and Arthur Hinsvark and Arun Rao and Aston Zhang and Aurelien Rodriguez and Austen Gregerson and Ava Spataru and Baptiste Roziere and Bethany Biron and Binh Tang and Bobbie Chern and Charlotte Caucheteux and Chaya Nayak and Chloe Bi and Chris Marra and Chris McConnell and Christian Keller and Christophe Touret and Chunyang Wu and Corinne Wong and Cristian Canton Ferrer and Cyrus Nikolaidis and Damien Allonsius and Daniel Song and Danielle Pintz and Danny Livshits and David Esiobu and Dhruv Choudhary and Dhruv Mahajan and Diego Garcia-Olano and Diego Perino and Dieuwke Hupkes and Egor Lakomkin and Ehab AlBadawy and Elina Lobanova and Emily Dinan and Eric Michael Smith and Filip Radenovic and Frank Zhang and Gabriel Synnaeve and Gabrielle Lee and Georgia Lewis Anderson and Graeme Nail and Gregoire Mialon and Guan Pang and Guillem Cucurell and Hailey Nguyen and Hannah Korevaar and Hu Xu and Hugo Touvron and Iliyan Zarov and Imanol Arrieta Ibarra and Isabel Kloumann and Ishan Misra and Ivan Evtimov and Jade Copet and Jaewon Lee and Jan Geffert and Jana Vranes and Jason Park and Jay Mahadeokar and Jeet Shah and Jelmer van der Linde and Jennifer Billock and Jenny Hong and Jenya Lee and Jeremy Fu and Jianfeng Chi and Jianyu Huang and Jiawen Liu and Jie Wang and Jiecao Yu and Joanna Bitton and Joe Spisak and Jongsoo Park and Joseph Rocca and Joshua Johnstun and Joshua Saxe and Junteng Jia and Kalyan Vasuden Alwala and Kartikeya Upasani and Kate Plawiak and Ke Li and Kenneth Heafield and Kevin Stone and Khalid El-Arini and Krithika Iyer and Kshitiz Malik and Kuenley Chiu and Kunal Bhalla and Lauren Rantala-Yeary and Laurens van der Maaten and Lawrence Chen and Liang Tan and Liz Jenkins and Louis Martin and Lovish Madaan and Lubo Malo and Lukas Blecher and Lukas Landzaat and Luke de Oliveira and Madeline Muzzi and Mahesh Pasupuleti and Mannat Singh and Manohar Paluri and Marcin Kardas and Mathew Oldham and Mathieu Rita and Maya Pavlova and Melanie Kambadur and Mike Lewis and Min Si and Mitesh Kumar Singh and Mona Hassan and Naman Goyal and Narjes Torabi and Nikolay Bashlykov and Nikolay Bogoychev and Niladri Chatterji and Olivier Duchenne and Onur Çelebi and Patrick Alrassy and Pengchuan Zhang and Pengwei Li and Petar Vasic and Peter Weng and Prajjwal Bhargava and Pratik Dubal and Praveen Krishnan and Punit Singh Koura and Puxin Xu and Qing He and Qingxiao Dong and Ragavan Srinivasan and Raj Ganapathy and Ramon Calderer and Ricardo Silveira Cabral and Robert Stojnic and Roberta Raileanu and Rohit Girdhar and Rohit Patel and Romain Sauvestre and Ronnie Polidoro and Roshan Sumbaly and Ross Taylor and Ruan Silva and Rui Hou and Rui Wang and Saghar Hosseini and Sahana Chennabasappa and Sanjay Singh and Sean Bell and Seohyun Sonia Kim and Sergey Edunov and Shaoliang Nie and Sharan Narang and Sharath Raparthy and Sheng Shen and Shengye Wan and Shruti Bhosale and Shun Zhang and Simon Vandenhende and Soumya Batra and Spencer Whitman and Sten Sootla and Stephane Collot and Suchin Gururangan and Sydney Borodinsky and Tamar Herman and Tara Fowler and Tarek Sheasha and Thomas Georgiou and Thomas Scialom and Tobias Speckbacher and Todor Mihaylov and Tong Xiao and Ujjwal Karn and Vedanuj Goswami and Vibhor Gupta and Vignesh Ramanathan and Viktor Kerkez and Vincent Gonguet and Virginie Do and Vish Vogeti and Vladan Petrovic and Weiwei Chu and Wenhan Xiong and Wenyin Fu and Whitney Meers and Xavier Martinet and Xiaodong Wang and Xiaoqing Ellen Tan and Xinfeng Xie and Xuchao Jia and Xuewei Wang and Yaelle Goldschlag and Yashesh Gaur and Yasmine Babaei and Yi Wen and Yiwen Song and Yuchen Zhang and Yue Li and Yuning Mao and Zacharie Delpierre Coudert and Zheng Yan and Zhengxing Chen and Zoe Papakipos and Aaditya Singh and Aaron Grattafiori and Abha Jain and Adam Kelsey and Adam Shajnfeld and Adithya Gangidi and Adolfo Victoria and Ahuva Goldstand and Ajay Menon and Ajay Sharma and Alex Boesenberg and Alex Vaughan and Alexei Baevski and Allie Feinstein and Amanda Kallet and Amit Sangani and Anam Yunus and Andrei Lupu and Andres Alvarado and Andrew Caples and Andrew Gu and Andrew Ho and Andrew Poulton and Andrew Ryan and Ankit Ramchandani and Annie Franco and Aparajita Saraf and Arkabandhu Chowdhury and Ashley Gabriel and Ashwin Bharambe and Assaf Eisenman and Azadeh Yazdan and Beau James and Ben Maurer and Benjamin Leonhardi and Bernie Huang and Beth Loyd and Beto De Paola and Bhargavi Paranjape and Bing Liu and Bo Wu and Boyu Ni and Braden Hancock and Bram Wasti and Brandon Spence and Brani Stojkovic and Brian Gamido and Britt Montalvo and Carl Parker and Carly Burton and Catalina Mejia and Changhan Wang and Changkyu Kim and Chao Zhou and Chester Hu and Ching-Hsiang Chu and Chris Cai and Chris Tindal and Christoph Feichtenhofer and Damon Civin and Dana Beaty and Daniel Kreymer and Daniel Li and Danny Wyatt and David Adkins and David Xu and Davide Testuggine and Delia David and Devi Parikh and Diana Liskovich and Didem Foss and Dingkang Wang and Duc Le and Dustin Holland and Edward Dowling and Eissa Jamil and Elaine Montgomery and Eleonora Presani and Emily Hahn and Emily Wood and Erik Brinkman and Esteban Arcaute and Evan Dunbar and Evan Smothers and Fei Sun and Felix Kreuk and Feng Tian and Firat Ozgenel and Francesco Caggioni and Francisco Guzmán and Frank Kanayet and Frank Seide and Gabriela Medina Florez and Gabriella Schwarz and Gada Badeer and Georgia Swee and Gil Halpern and Govind Thattai and Grant Herman and Grigory Sizov and Guangyi and Zhang and Guna Lakshminarayanan and Hamid Shojanazeri and Han Zou and Hannah Wang and Hanwen Zha and Haroun Habeeb and Harrison Rudolph and Helen Suk and Henry Aspegren and Hunter Goldman and Ibrahim Damlaj and Igor Molybog and Igor Tufanov and Irina-Elena Veliche and Itai Gat and Jake Weissman and James Geboski and James Kohli and Japhet Asher and Jean-Baptiste Gaya and Jeff Marcus and Jeff Tang and Jennifer Chan and Jenny Zhen and Jeremy Reizenstein and Jeremy Teboul and Jessica Zhong and Jian Jin and Jingyi Yang and Joe Cummings and Jon Carvill and Jon Shepard and Jonathan McPhie and Jonathan Torres and Josh Ginsburg and Junjie Wang and Kai Wu and Kam Hou U and Karan Saxena and Karthik Prasad and Kartikay Khandelwal and Katayoun Zand and Kathy Matosich and Kaushik Veeraraghavan and Kelly Michelena and Keqian Li and Kun Huang and Kunal Chawla and Kushal Lakhotia and Kyle Huang and Lailin Chen and Lakshya Garg and Lavender A and Leandro Silva and Lee Bell and Lei Zhang and Liangpeng Guo and Licheng Yu and Liron Moshkovich and Luca Wehrstedt and Madian Khabsa and Manav Avalani and Manish Bhatt and Maria Tsimpoukelli and Martynas Mankus and Matan Hasson and Matthew Lennie and Matthias Reso and Maxim Groshev and Maxim Naumov and Maya Lathi and Meghan Keneally and Michael L. Seltzer and Michal Valko and Michelle Restrepo and Mihir Patel and Mik Vyatskov and Mikayel Samvelyan and Mike Clark and Mike Macey and Mike Wang and Miquel Jubert Hermoso and Mo Metanat and Mohammad Rastegari and Munish Bansal and Nandhini Santhanam and Natascha Parks and Natasha White and Navyata Bawa and Nayan Singhal and Nick Egebo and Nicolas Usunier and Nikolay Pavlovich Laptev and Ning Dong and Ning Zhang and Norman Cheng and Oleg Chernoguz and Olivia Hart and Omkar Salpekar and Ozlem Kalinli and Parkin Kent and Parth Parekh and Paul Saab and Pavan Balaji and Pedro Rittner and Philip Bontrager and Pierre Roux and Piotr Dollar and Polina Zvyagina and Prashant Ratanchandani and Pritish Yuvraj and Qian Liang and Rachad Alao and Rachel Rodriguez and Rafi Ayub and Raghotham Murthy and Raghu Nayani and Rahul Mitra and Raymond Li and Rebekkah Hogan and Robin Battey and Rocky Wang and Rohan Maheswari and Russ Howes and Ruty Rinott and Sai Jayesh Bondu and Samyak Datta and Sara Chugh and Sara Hunt and Sargun Dhillon and Sasha Sidorov and Satadru Pan and Saurabh Verma and Seiji Yamamoto and Sharadh Ramaswamy and Shaun Lindsay and Shaun Lindsay and Sheng Feng and Shenghao Lin and Shengxin Cindy Zha and Shiva Shankar and Shuqiang Zhang and Shuqiang Zhang and Sinong Wang and Sneha Agarwal and Soji Sajuyigbe and Soumith Chintala and Stephanie Max and Stephen Chen and Steve Kehoe and Steve Satterfield and Sudarshan Govindaprasad and Sumit Gupta and Sungmin Cho and Sunny Virk and Suraj Subramanian and Sy Choudhury and Sydney Goldman and Tal Remez and Tamar Glaser and Tamara Best and Thilo Kohler and Thomas Robinson and Tianhe Li and Tianjun Zhang and Tim Matthews and Timothy Chou and Tzook Shaked and Varun Vontimitta and Victoria Ajayi and Victoria Montanez and Vijai Mohan and Vinay Satish Kumar and Vishal Mangla and Vítor Albiero and Vlad Ionescu and Vlad Poenaru and Vlad Tiberiu Mihailescu and Vladimir Ivanov and Wei Li and Wenchen Wang and Wenwen Jiang and Wes Bouaziz and Will Constable and Xiaocheng Tang and Xiaofang Wang and Xiaojian Wu and Xiaolan Wang and Xide Xia and Xilun Wu and Xinbo Gao and Yanjun Chen and Ye Hu and Ye Jia and Ye Qi and Yenda Li and Yilin Zhang and Ying Zhang and Yossi Adi and Youngjin Nam and Yu and Wang and Yuchen Hao and Yundi Qian and Yuzi He and Zach Rait and Zachary DeVito and Zef Rosnbrick and Zhaoduo Wen and Zhenyu Yang and Zhiwei Zhao},
      year={2024},
      eprint={2407.21783},
      archivePrefix={arXiv},
      primaryClass={cs.AI},
      url={https://arxiv.org/abs/2407.21783}, 
}

@article{Bai2024HallucinationOM,
  title={Hallucination of Multimodal Large Language Models: A Survey},
  author={Zechen Bai and Pichao Wang and Tianjun Xiao and Tong He and Zongbo Han and Zheng Zhang and Mike Zheng Shou},
  journal={ArXiv},
  year={2024},
  volume={abs/2404.18930},
  url={https://api.semanticscholar.org/CorpusID:269449935}
}

@inproceedings{rawte-etal-2023-troubling,
    title = "The Troubling Emergence of Hallucination in Large Language Models - An Extensive Definition, Quantification, and Prescriptive Remediations",
    author = "Rawte, Vipula  and
      Chakraborty, Swagata  and
      Pathak, Agnibh  and
      Sarkar, Anubhav  and
      Tonmoy, S.M Towhidul Islam  and
      Chadha, Aman  and
      Sheth, Amit  and
      Das, Amitava",
    editor = "Bouamor, Houda  and
      Pino, Juan  and
      Bali, Kalika",
    booktitle = "Proceedings of the 2023 Conference on Empirical Methods in Natural Language Processing",
    month = dec,
    year = "2023",
    address = "Singapore",
    publisher = "Association for Computational Linguistics",
    url = "https://aclanthology.org/2023.emnlp-main.155",
    doi = "10.18653/v1/2023.emnlp-main.155",
    pages = "2541--2573",
}

@inproceedings{wang-etal-2022-super,
    title = "Super-{N}atural{I}nstructions: Generalization via Declarative Instructions on 1600+ {NLP} Tasks",
    author = "Wang, Yizhong  and
      Mishra, Swaroop  and
      Alipoormolabashi, Pegah  and
      Kordi, Yeganeh  and
      Mirzaei, Amirreza  and
      Naik, Atharva  and
      Ashok, Arjun  and
      Dhanasekaran, Arut Selvan  and
      Arunkumar, Anjana  and
      Stap, David  and
      Pathak, Eshaan  and
      Karamanolakis, Giannis  and
      Lai, Haizhi  and
      Purohit, Ishan  and
      Mondal, Ishani  and
      Anderson, Jacob  and
      Kuznia, Kirby  and
      Doshi, Krima  and
      Pal, Kuntal Kumar  and
      Patel, Maitreya  and
      Moradshahi, Mehrad  and
      Parmar, Mihir  and
      Purohit, Mirali  and
      Varshney, Neeraj  and
      Kaza, Phani Rohitha  and
      Verma, Pulkit  and
      Puri, Ravsehaj Singh  and
      Karia, Rushang  and
      Doshi, Savan  and
      Sampat, Shailaja Keyur  and
      Mishra, Siddhartha  and
      Reddy A, Sujan  and
      Patro, Sumanta  and
      Dixit, Tanay  and
      Shen, Xudong",
    editor = "Goldberg, Yoav  and
      Kozareva, Zornitsa  and
      Zhang, Yue",
    booktitle = "Proceedings of the 2022 Conference on Empirical Methods in Natural Language Processing",
    month = dec,
    year = "2022",
    address = "Abu Dhabi, United Arab Emirates",
    publisher = "Association for Computational Linguistics",
    url = "https://aclanthology.org/2022.emnlp-main.340/",
    doi = "10.18653/v1/2022.emnlp-main.340",
    pages = "5085--5109"
}

@misc{chung2022scalinginstructionfinetunedlanguagemodels,
      title={Scaling Instruction-Finetuned Language Models}, 
      author={Hyung Won Chung and Le Hou and Shayne Longpre and Barret Zoph and Yi Tay and William Fedus and Yunxuan Li and Xuezhi Wang and Mostafa Dehghani and Siddhartha Brahma and Albert Webson and Shixiang Shane Gu and Zhuyun Dai and Mirac Suzgun and Xinyun Chen and Aakanksha Chowdhery and Alex Castro-Ros and Marie Pellat and Kevin Robinson and Dasha Valter and Sharan Narang and Gaurav Mishra and Adams Yu and Vincent Zhao and Yanping Huang and Andrew Dai and Hongkun Yu and Slav Petrov and Ed H. Chi and Jeff Dean and Jacob Devlin and Adam Roberts and Denny Zhou and Quoc V. Le and Jason Wei},
      year={2022},
      eprint={2210.11416},
      archivePrefix={arXiv},
      archivePrefix={arXiv},
      primaryClass={cs.LG},
      url={https://arxiv.org/abs/2210.11416}, 
}

@article{Farquhar2024,
  title = {Detecting hallucinations in large language models using semantic entropy},
  volume = {630},
  ISSN = {1476-4687},
  url = {http://dx.doi.org/10.1038/s41586-024-07421-0},
  DOI = {10.1038/s41586-024-07421-0},
  number = {8017},
  journal = {Nature},
  publisher = {Springer Science and Business Media LLC},
  author = {Farquhar,  Sebastian and Kossen,  Jannik and Kuhn,  Lorenz and Gal,  Yarin},
  year = {2024},
  month = jun,
  pages = {625–630}
}

@misc{kadavath2022languagemodelsmostlyknow,
      title={Language Models (Mostly) Know What They Know}, 
      author={Saurav Kadavath and Tom Conerly and Amanda Askell and Tom Henighan and Dawn Drain and Ethan Perez and Nicholas Schiefer and Zac Hatfield-Dodds and Nova DasSarma and Eli Tran-Johnson and Scott Johnston and Sheer El-Showk and Andy Jones and Nelson Elhage and Tristan Hume and Anna Chen and Yuntao Bai and Sam Bowman and Stanislav Fort and Deep Ganguli and Danny Hernandez and Josh Jacobson and Jackson Kernion and Shauna Kravec and Liane Lovitt and Kamal Ndousse and Catherine Olsson and Sam Ringer and Dario Amodei and Tom Brown and Jack Clark and Nicholas Joseph and Ben Mann and Sam McCandlish and Chris Olah and Jared Kaplan},
      year={2022},
      eprint={2207.05221},
      archivePrefix={arXiv},
      primaryClass={cs.CL},
      url={https://arxiv.org/abs/2207.05221}, 
}

@article{He-Deberta-2020,
  author       = {Pengcheng He and
                  Xiaodong Liu and
                  Jianfeng Gao and
                  Weizhu Chen},
  title        = {DeBERTa: Decoding-enhanced {BERT} with Disentangled Attention},
  journal      = {CoRR},
  volume       = {abs/2006.03654},
  year         = {2020},
  url          = {https://arxiv.org/abs/2006.03654},
  eprinttype    = {arXiv},
  eprint       = {2006.03654},
  bibsource    = {dblp computer science bibliography, https://dblp.org}
}

@article{mmlu,
  author       = {Dan Hendrycks and
                  Collin Burns and
                  Steven Basart and
                  Andy Zou and
                  Mantas Mazeika and
                  Dawn Song and
                  Jacob Steinhardt},
  title        = {Measuring Massive Multitask Language Understanding},
  journal      = {CoRR},
  volume       = {abs/2009.03300},
  year         = {2020},
  url          = {https://arxiv.org/abs/2009.03300},
  eprinttype    = {arXiv},
  eprint       = {2009.03300},
  bibsource    = {dblp computer science bibliography, https://dblp.org}
}

@article{siqa,
  author       = {Maarten Sap and
                  Hannah Rashkin and
                  Derek Chen and
                  Ronan Le Bras and
                  Yejin Choi},
  title        = {SocialIQA: Commonsense Reasoning about Social Interactions},
  journal      = {CoRR},
  volume       = {abs/1904.09728},
  year         = {2019},
  url          = {http://arxiv.org/abs/1904.09728},
  eprinttype    = {arXiv},
  eprint       = {1904.09728},
  bibsource    = {dblp computer science bibliography, https://dblp.org}
}

@article{hu-etal-lora-2021,
  author       = {Edward J. Hu and
                  Yelong Shen and
                  Phillip Wallis and
                  Zeyuan Allen{-}Zhu and
                  Yuanzhi Li and
                  Shean Wang and
                  Weizhu Chen},
  title        = {LoRA: Low-Rank Adaptation of Large Language Models},
  journal      = {CoRR},
  volume       = {abs/2106.09685},
  year         = {2021},
  url          = {https://arxiv.org/abs/2106.09685},
  eprinttype    = {arXiv},
  eprint       = {2106.09685},
  bibsource    = {dblp computer science bibliography, https://dblp.org}
}

@inproceedings{
ramasesh2022effect,
title={Effect of scale on catastrophic forgetting in neural networks},
author={Vinay Venkatesh Ramasesh and Aitor Lewkowycz and Ethan Dyer},
booktitle={International Conference on Learning Representations},
year={2022},
url={https://openreview.net/forum?id=GhVS8_yPeEa}
}

@article{gsm8k,
  author       = {Karl Cobbe and
                  Vineet Kosaraju and
                  Mohammad Bavarian and
                  Mark Chen and
                  Heewoo Jun and
                  Lukasz Kaiser and
                  Matthias Plappert and
                  Jerry Tworek and
                  Jacob Hilton and
                  Reiichiro Nakano and
                  Christopher Hesse and
                  John Schulman},
  title        = {Training Verifiers to Solve Math Word Problems},
  journal      = {CoRR},
  volume       = {abs/2110.14168},
  year         = {2021},
  url          = {https://arxiv.org/abs/2110.14168},
  eprinttype    = {arXiv},
  eprint       = {2110.14168},
  bibsource    = {dblp computer science bibliography, https://dblp.org}
}

@misc{wang2024mmluprorobustchallengingmultitask,
      title={MMLU-Pro: A More Robust and Challenging Multi-Task Language Understanding Benchmark}, 
      author={Yubo Wang and Xueguang Ma and Ge Zhang and Yuansheng Ni and Abhranil Chandra and Shiguang Guo and Weiming Ren and Aaran Arulraj and Xuan He and Ziyan Jiang and Tianle Li and Max Ku and Kai Wang and Alex Zhuang and Rongqi Fan and Xiang Yue and Wenhu Chen},
      year={2024},
      eprint={2406.01574},
      archivePrefix={arXiv},
      primaryClass={cs.CL},
      url={https://arxiv.org/abs/2406.01574}, 
}
\bibliographystyle{iclr2026_conference}
\appendix
\section{Discussing semantic representations}\label{sec:semantic}\label{sec:open-experiments}
In this paper, we \textcolor{black}{generally} focus on the relatively easy task of consolidating semantically similar answers for multiple-choice question answering datasets. In this case, semantic normalization is trivial, as it simply requires isolating the letter of the multiple-choice option, removing the reasoning and punctuation that affect lexical uncertainty quantification methods. However,
for more complex tasks other approaches may be required~\citep{huang2024calibrating}. Previous research has established how normalization might be applied: for example,~\citet{kuhn2023semantic} use natural language inference to cluster semantically equivalent answers and \citet{tian-etal-2023-just} use an LLM as a judge of correctness.
\color{black}
To demonstrate this variant, we set up an experiment to demonstrate how uncertainty distillation can be applied to an open-answer dataset.
\subsection{Open-answer experiments}
\paragraph{Model and dataset} We run these experiments with Llama-3B-Instruct. For the open dataset, we use GSM8K \citep{gsm8k}, an open math QA dataset consisting of grade-school level math problems. This dataset presents input variance that prevent exact match metrics from working effectively: even assuming the model correctly only encloses the final answer in the tags, an answer might be expressed as ``10'', ``10 dollars'', ``\$10'', ``10.00'', and so on. All of these answers are semantically equivalent, but ``10'' would be the only accepted answer. We take the first 7000 examples of the training set as training data, the remaining 473 examples as validation data, and the existing test set as the unseen test set.

\paragraph{Semantic normalization} To make the clusters, we use code from \citet{kuhn2023semantic}, specifically the EntailmentDeberta with minor changes to look for the absence of contradiction rather than entailment\footnote{As Deberta~\citep{He-Deberta-2020} is trained for natural language inference, rather than comparing two numbers, absence of contradiction works better to cluster than entailment.}. Once each sample has been generated, we compare answers pairwise, first to the correct answer (Formatted as ``The correct answer is'' followed by the simple numerical answer), and then to existing clusters. If none match, the answer is assigned to a new cluster. We choose a random answer to represent each cluster when constructing training data. The remainder of uncertainty distillation proceeds as normal.

\paragraph{Baselines and metrics} The baselines are described in \autoref{sec:experiments}. For P(IK), rather than using exact match to assign correctness labels to train the probe, we use EntailmentDeberta. At inference, to evaluate generated answers for all baselines, we query GPT-3.5-turbo as a judge.

\paragraph{Results and analysis}
We find that uncertainty distillation in this setting outperforms all baselines by a wide margin, achieving AUROC of 0.787. Both AUROC and high accuracy are significantly higher than the two baselines we compare to, and AUROC is similarly high to our multiple-choice question answering results, demonstrating that uncertainty distillation can be successfully applied to open-answer tasks by using semantic clustering to normalize answers at data generation.
\begin{table*}[h]
    \centering
    \begin{small}
    \begin{sc}
    \begin{tabular}{c|cc|cc}
    \toprule
          \color{black}Method &\color{black}AUROC& \color{black}Acc& \color{black}High Acc &\color{black}High \%    \\
         \midrule
          \color{black}UD (ours) & \color{black}\textbf{0.787}&\color{black}0.752& \color{black}0.935&\color{black}58.0\\
          \color{black}Lexical baseline&\color{black}0.542& \color{black}0.829 &\color{black} 0.832  &\color{black} 98.2 \\
          \color{black}Prompting&\color{black}0.587& \color{black}0.763&\color{black}0.803 & \color{black}63.5\\
          %\color{black}P(IK) & &&&\\
         % P(True)&&&&\\
         % Sem. Entropy & & &&\\
          \bottomrule

    \end{tabular}
    \end{sc}
    \end{small}
    \caption{\textcolor{black}{AUROC and accuracy metrics for the open-answer experiments with Llama-3B-Instruct. We find that uncertainty distillation (UD) leads to increased AUROC and accuracy in high-confidence categories.}}
    \label{tab:open}
\end{table*}

\subsection{Generalization to long-form tasks}\label{sec:summarization}

For longer-form generation tasks, a single binary confidence judgment may be inadequate. The uncertainty distillation procedure is straightforward to extend to settings involving more than one prediction per generation, providing more granular feedback at the level of individual claims. As a concrete example, consider the task of extracting key facts from a news article. To apply our framework to this setting, we use a structured output consisting of a set of independent predictions, each of which associated with a \emph{different} confidence. For a document from the CNN/DailyMail corpus\footnote{\url{abisee/cnn_dailymail}}, we show how \texttt{gemini-2.5-flash-lite} extracts key facts out-of-the-box using a suitable prompt and JSON output. However, as shown in~\autoref{sec:api-tuning}, the calibration of the verbalized confidences is poor. By sampling multiple sets of claims for the same document, we can estimate the probability of each fact independently, and then apply our calibration and data augmentation procedure to fine-tune a model that outputs better calibrated confidences without incurring any additional runtime overhead. This requires using a semantic clustering approach which groups together equivalent factual statements expressed differently, which may be the same base model or a more efficient natural language inference model~\citep{kuhn2023semantic}.

\textbf{Prompt}: Read the following document and summarize at most 5 key facts of the story as a list of claims (fewer is fine). Keep the claims short, simple, and in canonical format. Return your answer in JSON format with the field 'claims', which is a list of objects containing 'text' and 'confidence'.

\textbf{Document}: Solar flares can cause widespread disruption on Earth, but scientists have struggled to understand how they are formed. Now, a team from from the California Institute of Technology (Caltech) have taken a huge step forward by recreated the phenomenon that leads to solar flares in a lab. They hope their work could lead to improved space weather forecasts. The US team amanged to recreate plasma loops (above) in a lab for the first time [...]
\begin{lstlisting}[language=json,firstnumber=1]
{
  "claims": [
    {
      "text": "Scientists have recreated the phenomenon that leads to solar flares in a lab.",
      "confidence": "VERY HIGH"
    },
    {
      "text": "The laboratory recreation involved creating plasma loops, which are precursors to solar flares.",
      "confidence": "VERY HIGH"
    },
    {
      "text": "The Caltech team hopes this research will lead to improved space weather forecasts and warnings.",
      "confidence": "VERY HIGH"
    },
    {
      "text": "Two magnetic forces were identified as controlling the behavior of the arching plasma loops.",
      "confidence": "HIGH"
    },
    {
      "text": "A novel technique using color-coded, inflowing plasma was employed to visualize the plasma loop behavior.",
      "confidence": "HIGH"
    }
  ]
}
\end{lstlisting}
\color{black}
\section{Targeted analysis}\label{sec:inst-tuning}
\subsection{Experimental setting}
In \autoref{sec:method}, we assume that we have access to held-out calibration data. However, due to the unknown composition and size of pretraining datasets, it is increasingly challenging to guarantee that this assumption holds. We therefore test uncertainty distillation in a setting where we can know with certainty whether the calibration set is in a model's pretraining data. We examine this question and the impact of adding varying numbers of incorrect answers during uncertainty distillation in \autoref{sec:inst-tuning}.

\paragraph{Dataset} 
In this setting, we use the Super-NaturalInstructions dataset~\citep[SNI;][]{wang-etal-2022-super}.
We select $15$ English Q\&A tasks with short-form answers.
We focus on Q\&A tasks for which a single correct answer exists (e.g. multiple choice problems, short-form span extraction, math problems, etc.) and thus for which correctness of a model's prediction can reliably and efficiently be computed after normalizing lexical forms without resorting to methods such as LLM verification. We use 1,000 samples to obtain our Monte Carlo estimate of confidence (see \autoref{sec:samples} for details on how number of samples affects successful confidence estimation).

\paragraph{Models} 
 We perform uncertainty distillation on FLAN-T5 \citep{chung2022scalinginstructionfinetunedlanguagemodels}, an instruction-tuned model trained on a dataset containing the SNI tasks.
Importantly, we not only verify that Flan-T5 has been instruction-tuned on our tasks, but has seen samples from the {\it calibration set} of our test tasks.
This allows us to investigate the effect of data contamination on calibration of verbalized confidences. 

To construct a similar model which has \textit{not} seen our calibration data, we instruction-tune a T5-Large model on a remaining subset of the English tasks in the SNI dataset, making sure to explicitly hold out the $15$ tasks we use in our uncertainty distillation experiments.
The result is an instruction-tuned model which we refer to as Instruct-T5, capable of performing our target Q\&A tasks without having seen these tasks during training. In other words, the samples we obtain from this model do not require Instruct-T5 to be pre-trained on that specific task.
See \autoref{app:sni-data} for more details on our data selection and instruction-tuning. We train and evaluate \methodName on the combined dataset of these tasks and report the performance over the metrics described in \autoref{sec:metrics}.

\paragraph{Baselines} We report a comparison to the lexical baseline described above in order to provide context for the performance of the small models. 
\subsection{Results}

\paragraph{Assumption of calibration set}\label{sec:flan} We compare the performance of FLAN-T5, which has been instruction-tuned on the calibration set, with the performance of Instruct-T5, which has not, in  \autoref{tab:flan}.
We find that while \methodName still produces meaningful confidence bins for FLAN-T5, it no longer outperforms lexical uncertainty. We conclude that uncertainty distillation works in the absence of held-out calibration data, but not as effectively as token-level probabilities, which are likely well-calibrated due to the model's previous training on these examples. We discuss results for these two models further in  \autoref{sec:incorrect} and \autoref{sec:post-hoc}, and find that the behavior of FLAN-T5 differs significantly from results on models where we have an unseen calibration set.
\begin{table*}[h]
    \centering
    \begin{small}
    \begin{sc}
    \begin{tabular}{c|c|cc|c}
    \toprule
          Model & Method &AUROC& Overall Accuracy& High Accuracy   \\
            \midrule
          \multirow{2}{*}{Instruct-T5} &
           Uncertainty distillation & \textbf{0.751} & \textbf{0.449} & \textbf{0.839}\\
          &Lexical baseline&0.667 &0.387&0.754\\
         \midrule
          \multirow{2}{*}{FLAN-T5} &Uncertainty distillation &0.873& 0.614 &0.875 \\
          &Lexical baseline&\textbf{0.892}& \textbf{0.657}& \textbf{0.912} \\
         \bottomrule

    \end{tabular}
    \end{sc}
    \end{small}
    \caption{AUROC and accuracy metrics when using FLAN-T5, which does not have an unseen calibration set. We find that while uncertainty distillation outperforms our lexical baseline with a model with an unseen calibration set, it does not outperform the baseline on FLAN-T5, which was instruction-tuned on the data previously.}
    \label{tab:flan}
\end{table*}
\paragraph{Adding incorrect examples}\label{sec:incorrect} While adding incorrect examples into the training data has the potential to provide more examples at different levels of confidences, it also is likely to increase the likelihood that a model generates an incorrect answer. To demonstrate this effect, in \autoref{tab:incorrect}, we show the AUROC and accuracy for models trained with different amounts of incorrect samples. With Instruct-T5, we find that adding only two incorrect samples per correct sample dramatically increases AUROC while decreasing accuracy. While this would seem to indicate a fundamental tradeoff between accuracy and calibration, we find that the same is not as obviously true for FLAN-T5; while the accuracy may decrease and AUROC may increase, the effects are not as significant as they are for Instruct-T5. One possible interpretation of this is that its predictions are shaped by the fact that the data was included in its instruction-tuning corpus, leading to less dramatic shifts when trained. 

 While adding incorrect samples may improve AUROC, it increases the number of training examples by a factor of the number of incorrect examples added (e.g. a training set with 100 examples would train on 100 augmented answers with 0 incorrect examples added, 200 augmented answers with one incorrect example added, etc.) This leads to increased compute at training time. For this reason, in addition to the decreased accuracy, we recommend adding a low number of incorrect examples to the training dataset, and in our main experiments limit to at most one incorrect answer per question.
\begin{table}[h]
    \centering
    \begin{small}
    \begin{sc}
    \begin{tabular}{c|cccc}
    \toprule
          & \textbf{0} & \textbf{1} & \textbf{2} & \textbf{3} \\
         \midrule
       %   \multicolumn{5}{c}{T5-Base} \\
       %   \midrule
       %     AUROC  &  0.707& 0.755 &0.805 & \textbf{0.838} \\
       % Accuracy  &  \textbf{0.791}& 0.755&0.711& 0.652 \\
       %          \midrule 
          \multicolumn{5}{c} {Instruct-T5} \\
         \midrule
           AUROC  &  0.723& 0.737&0.751 & \textbf{0.757} \\
       Accuracy  &  \textbf{0.529}& 0.486 &0.449& 0.447 \\
                \midrule 
               \multicolumn{5}{c} {FLAN-T5} \\
         \midrule
           AUROC  &  0.868& 0.876& 0.873 & \textbf{0.883} \\
       Accuracy  &  0.609& \textbf{0.620}&0.614& 0.611 \\
                \bottomrule

    \end{tabular}
    \end{sc}
    \end{small}
    \caption{AUROC of models trained with varying numbers of incorrect examples allowed per question. There is a general trend towards increasing AUROC and decreasing accuracy when incorrect examples are included, although this is less pronounced for FLAN-T5. }
    \label{tab:incorrect}
\end{table}
\subsection{Analysis}
One high-level takeaway is that with small models there appears to be a tradeoff between an LLM's ability to predict its own confidence and overall model accuracy, but that this effect is less obvious with increasing model sizes. In our small-scale analysis, interventions that improve AUROC decrease accuracy and vice versa; however, with larger models we do not note as noticeable a decrease in accuracy compared to our baselines. \textcolor{black}{Functionally, UD combines aspects of two tasks: the model's original question answering ability and uncertainty quantification. Large models are both less prone to catastrophic forgetting\citep{ramasesh2022effect}and more effective at multitask learning than smaller models\citep{chung2022scalinginstructionfinetunedlanguagemodels}. With this framing, the fact that larger models' accuracy is less impaired by the finetuning process of uncertainty distillation indicates that model scale plays a significant role in an accuracy/performance tradeoff, and increasing model scale or training in an explicitly multi-task setting may decrease the likelihood of drops in accuracy.}
\section{Uncertainty distillation on supervised fine-tuned models}\label{sec:squad}
We here examine uncertainty distillation's efficacy when performed on a small fine-tuned model, rather than large instruction-tuned models.
\paragraph{Dataset} We perform these experiments using the SQuAD benchmark~\citep{rajpurkar2016squad}. This is a machine-reading task where each question consists of a passage of text and one or more associated questions, each of which is answerable based on the text itself. As the test set has not been publicly released, we use the splits proposed by \citet{DBLP:journals/corr/DuSC17}, which divides the publicly available available training and validation splits into train, test, and validation splits. We consider the first 60,000 examples in the training set to be training data, and the remainder to be our calibration set. %The benchmark was subsequently expanded to included non-answerable questions~\citep{rajpurkar2018know}, although we focus on the initial version in our current experiments.

\paragraph{Model} We apply uncertainty distillation to T5-base~\citep{raffel2020exploring} finetuned on a portion of SQUAD. We use defaults for most hyperparameters, and report hyperparameters in \autoref{sec:hparams}.

\paragraph{Results}
 \autoref{tab:squad} shows the results on the fine-tuned T5-base model. Uncertainty distillation achieves AUROC of 0.805 in the T5-base SQUAD experiment, slightly outperforming the lexical baseline's AUROC of 0.771.
\begin{table*}[h]
    \centering
    \begin{small}
    \begin{sc}
    \begin{tabular}{c|c|cc|c}
        \toprule
          Model & Method &AUROC & Overall Accuracy& High Accuracy   \\
         \midrule
         \multirow{2}{*}{T5-base}&
          Uncertainty distillation&0.805  &0.711&0.852 \\
          &Lexical baseline &0.771& 0.811& 0.865 \\
          \bottomrule
         
    %\end{sc}     
    \end{tabular}
    \end{sc}
    \end{small}
    \caption{AUROC and accuracy metrics for T5-base, trained on SQUAD. We find that even in this setting, a model trained with uncertainty distillation outperforms lexical uncertainty in verbalizing confidences on SQUAD-T5}
    \label{tab:squad}
\end{table*}
\section{Number of samples}\label{sec:samples}
\begin{figure}[h]
\centering
\includegraphics[scale=.35]{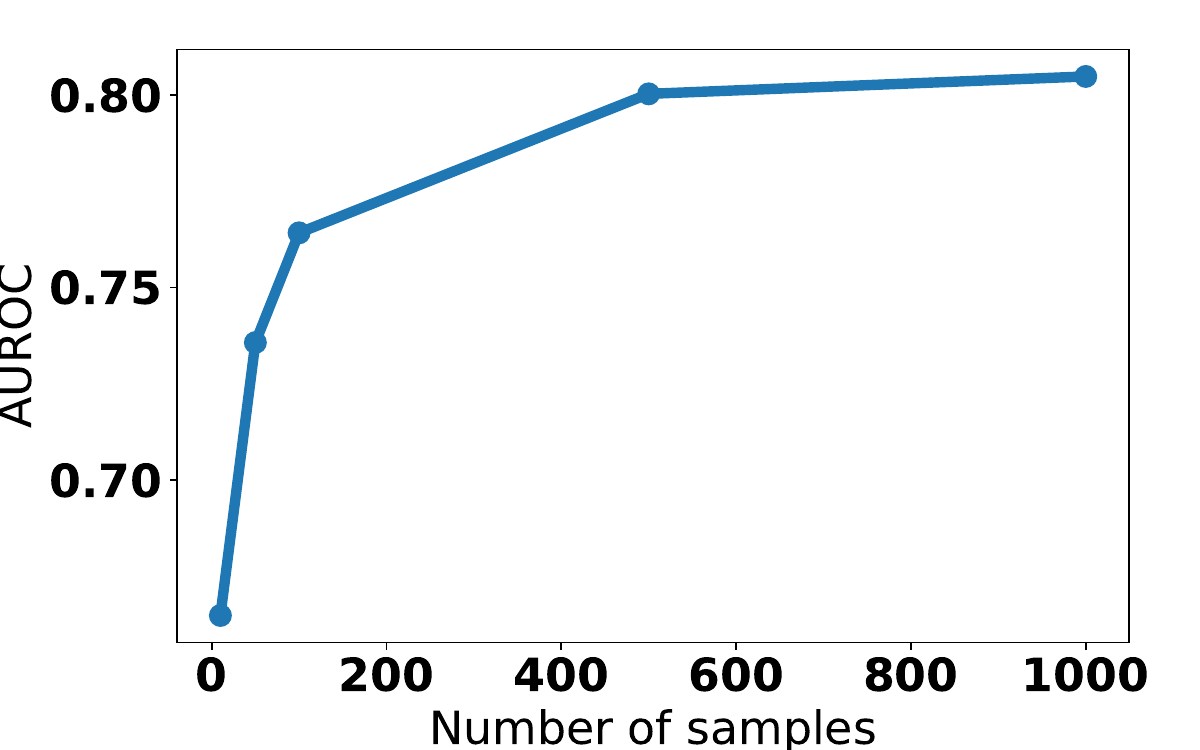}
  \caption{Curve showing the AUROC as a function of number of samples on the SQUAD dataset.}\label{fig:samples}
\end{figure}

Our Monte Carlo estimation of probability requires sampling repeatedly from a model before normalizing and calculating probability. In \autoref{fig:samples}, we show that the number of samples used to estimate the initial probabilities has a significant impact if chosen to be too low; however, there are diminishing returns as the number of samples increases. We therefore choose to use 1,000 samples in all of our experiments with FLAN-T5 and Instruct-T5, as more than that is unlikely to achieve anything but marginal improvement. \textcolor{black}{For the larger models, we select 100 samples, as this appears to be the elbow of the curve in \autoref{fig:samples}, and as sampling 1000 samples from the large models would be computationally prohibitive. We note, however, that based on these results, this hyperparameter can be changed to improve efficiency or effectiveness of the method as is required by each task.}
\section{Prompts}\label{sec:prompts}
\subsection{Mistral, Llama}
\paragraph{Prompt baselines, uncertainty distillation (MC)} \texttt{Answer the following question and state confidence in the answer (very low, low, medium, high, very high). Enclose concise reasoning in <reasoning> </reasoning> tags, confidence in <confidence> </confidence> tags, and the letter of your FINAL answer in <answer> </answer> tags without any of your work, like this: "If each of Lisa's 7 chickens lays 6 eggs, how many eggs does Lisa have? \\ A) 24 \\ B) 35 \\ C) 42 \\ D) 50 \\<reasoning> This can be solved with multiplication. The answer is 7*6, or 42.</reasoning> <answer> C) 42 </answer> <confidence>very high</confidence>." Your answer should not include words.}
\paragraph{Prompt baseline, uncertainty distillation (open)}
\textcolor{black}{\texttt{"You are a helpful AI assistant. Answer the following math question as briefly as possible and accurately. Enclose confidence in the answer (very low, low, medium, high, very high) after the answer in <confidence> </confidence> tags, like so: <confidence> very high </confidence>."}}
\paragraph{Sampling, lexical baseline} \texttt{Answer the following question. Enclose concise reasoning in <reasoning> </reasoning> tags and the letter of your FINAL answer in <answer> </answer> tags without any of your work, like this: "If each of Lisa's 7 chickens lays 6 eggs, how many eggs does Lisa have? \\
A) 24 \\
B) 35\\
C) 42\\
D) 50 \\
<reasoning> This can be solved with multiplication. The answer is 7*6, or 42.</reasoning> <answer> C) 42 </answer>." Your answer should not include words.}
\paragraph{Sampling, lexical baseline (open)}
\textcolor{black}{\texttt{"You are a helpful AI assistant. Answer the following math question as briefly as possible and accurately."}}
\subsection{Instruct-T5, FLAN-T5}
Each task in SNI has an associated instruction. For \textbf{sampling and the lexical baseline}, we simply use this instruction. For uncertainty distillation, we append \texttt{``Additionally state how confident you are in your answer''} to the instruction.
\subsection{\textcolor{black}{LLM-as-a-judge}}
\textcolor{black}{\texttt{We are evaluating answers to the question \"\{question\}\"\\
Here are two possible answers:\\
        Possible Answer 1: \{text1\}\\
        Possible Answer 2: \{text2\}   \\  
        Is Possible Answer 1 equivalent to Possible Answer 2, or do the answers contradict? Respond only with 'equivalent' or 'contradictory'.}}
\section{\textcolor{black}{Bin and label choice}}\label{sec:binning}
\begin{figure*}[t]
\centering
\includegraphics[scale=.22]{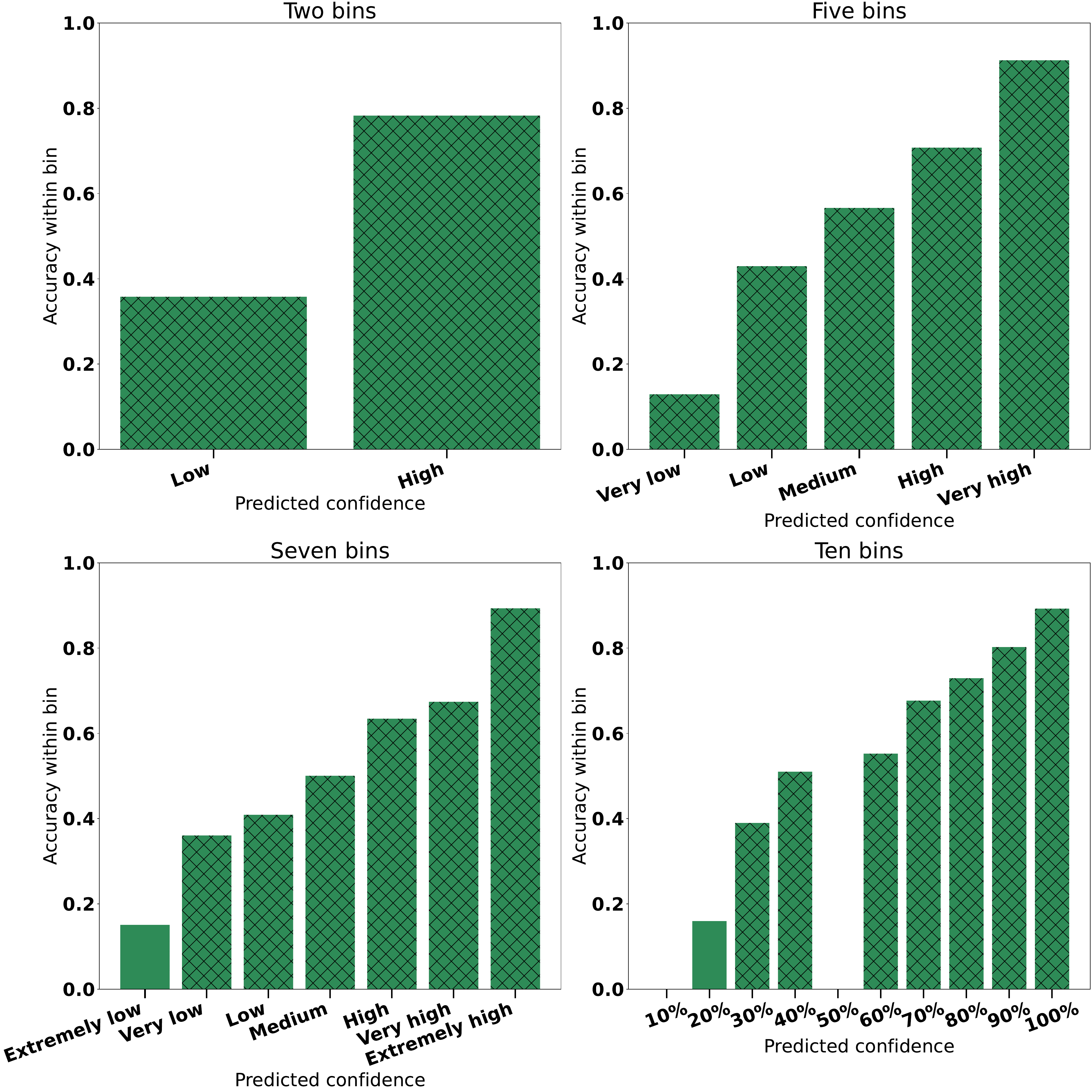}
  \caption{\textcolor{black}{Running UD on SIQA with Llama-3B and changing the number of the bins, or the labeling scheme, has no noticeable effect on the efficacy of UD aside from increased sparsity in bins. As in other figures, bins with fewer than 10 samples are not plotted.}}\label{fig:bin_size}
\end{figure*}

\textcolor{black}{In the main experiments, we examine the effect of UD with five bins and a verbalized naming scheme. However,  in \autoref{fig:bin_size}, we examine the effect of running UD on SocialIQA with Llama-3B while varying the number of bins (and thus necessarily changing the labeling scheme). Here, we find appropriate calibration regardless of number of bins. Notably, even changing the labeling scheme to numerical percentages does not result in a change in performance, suggesting that UD is robust to variance in labeling schemes. We use five bins as the default, as it offers enough bins to be challenging while avoiding the problems of sparsity (and thus noisiness) in bins that arise with larger bin sizes.}

\section{Effects of post-hoc calibration}\label{sec:post-hoc}
If the model's initial predictions are poorly calibrated, the post-hoc calibration step should help to better align probabilities in the training data with the true likelihood of success. \textcolor{black}{In \autoref{tab:posthoc-large}, we compare the miscalibration of the training data (measured through ECE with 30 bins) to the performance of models with and without post-hoc calibration during data generation. Unsurprisingly, we find that post-hoc calibration has positive effect corresponding to the initial miscalibration of the training data. For instance, Llama-3B on SocialIQA achieves 0.784 AUROC when trained on post-hoc calibrated data, and only 0.691 when identically trained on data without post-hoc calibration, with an ECE of 0.10. However, Ministral-8B on SocialIQA has a comparatively small ECE of 0.026, and the performance without post-hoc calibration is equivalent to the performance with post-hoc calibration. We conclude that the decision to include post-hoc calibration can be quickly and cheaply made by simply measuring the calibration of the annotated training data.}

\begin{table*}[h]
    \centering
    \begin{small}
    \begin{sc}
    \begin{tabular}{c|c|cc|c}
    
    \toprule
          \color{black}Dataset &\color{black}Model & \color{black}With Post-hoc & \color{black}No Post-hoc& \color{black}Training Data ECE \\
         \midrule
           \multirow{2}{*}{\color{black}MMLU} &\color{black}Ministral-8B & \color{black}\textbf{0.693} & \color{black}0.689 & \color{black}0.033\\
           &\color{black}Llama-3B & \color{black}\textbf{0.743} & \color{black}0.714 &\color{black} 0.039\\
                \hline 
           \multirow{2}{*}{\color{black}SIQA}&\color{black}Ministral-8B & \color{black}0.671& \color{black}\textbf{0.673} & \color{black}0.026\\
           &\color{black}Llama-3B & \color{black}\textbf{0.784}& \color{black}0.691 & \color{black}0.100\\
           \bottomrule

    \end{tabular}
    \end{sc}
    \end{small}
    \caption{\textcolor{black}{AUROC of large models with and without post-hoc calibration at training time. We find that post-hoc calibration tends to improve performance, most dramatically with Llama-3B on SIQA.}}
    \color{black}
    \label{tab:posthoc-large}
\end{table*}

\begin{figure*}[h]
\centering
\includegraphics[scale=.3]{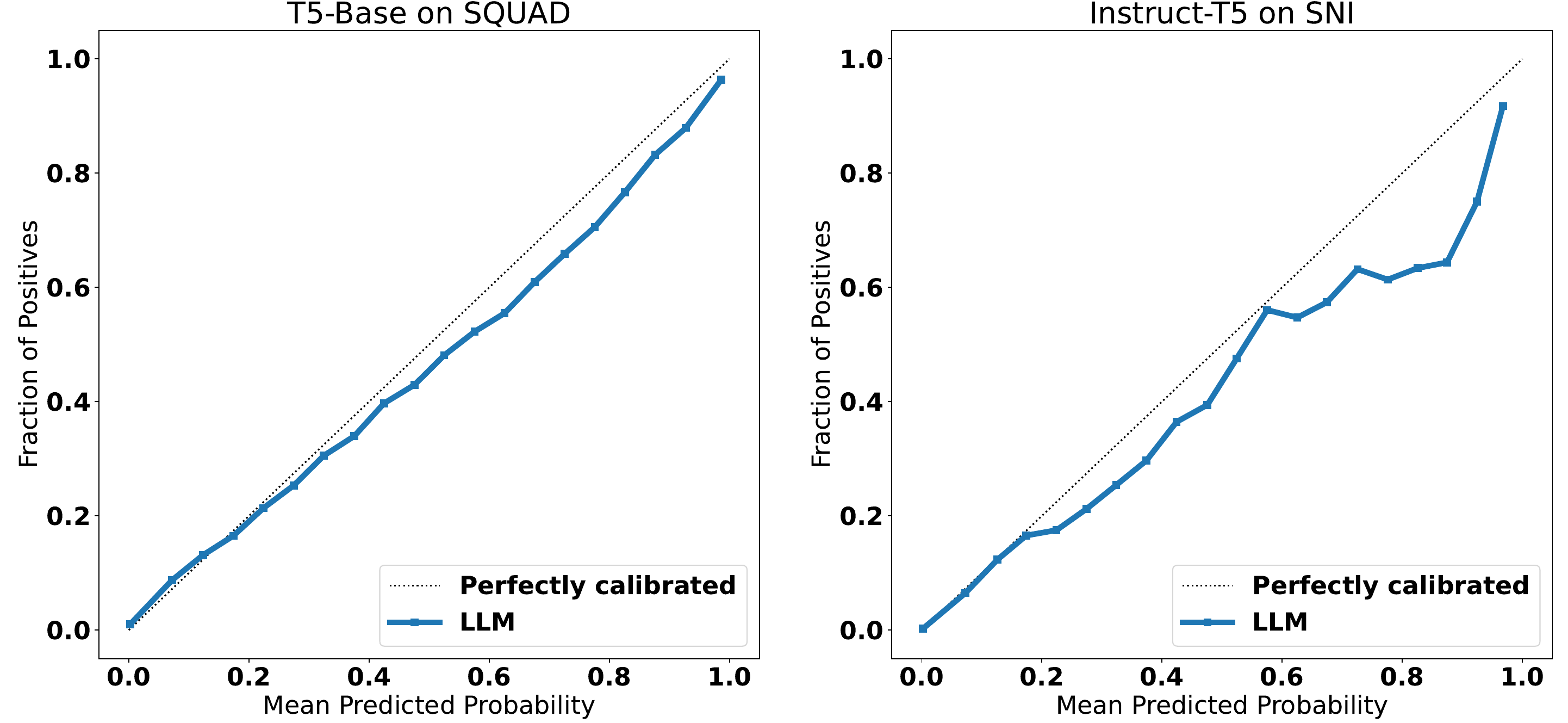}
  \caption{Initial calibration of our T5-base and Instruct-T5 model. Both models are well-calibrated in their respective domains, indicating that post-hoc calibration may not be necessary.}\label{fig:calibration}
\end{figure*}

\textcolor{black}{We further analyze how} post-hoc calibration impact\textcolor{black}{s} the model when \textcolor{black}{small models are} already well-calibrated on the specific task. \autoref{fig:calibration} shows the reliability diagrams for T5-base on SQUAD and Instruct-T5 on SNI. The models' predicted confidences align well with their actual accuracies; this allows us to investigate whether post-hoc calibration has a significant impact on AUROC \textcolor{black}{for smaller models}. Additionally, FLAN-T5 has been previously tuned on our calibration set; this gives us a setting to investigate the impact of post-hoc calibration when unseen calibration data is unavailable \textcolor{black}{and the model is presumably correctly confident in its predictions}. 

In \autoref{tab:posthoc}, we show the results of the smaller models trained with and without this post-hoc calibration step. We find no apparent benefit of post-hoc calibration for Instruct-T5 or fine-tuned T5-base. These models are already well-calibrated on their domains; \textcolor{black}{similarly to large models}, a post-hoc calibrator does not significantly alter the output probabilities. 

In the case of FLAN-T5, post-hoc calibration decreases AUROC. This suggests that in cases when unseen calibration data cannot be obtained \textcolor{black}{for small models}, \methodName may be more effective without the post-hoc calibration step.

\begin{table*}[h]
    \centering
    \begin{small}
    \begin{sc}
    \begin{tabular}{c|c|cc}
    \toprule
          Dataset &Model & With Post-hoc & No Post-hoc \\
         \midrule
           SQUAD &T5-base & 0.804 & 0.800 \\
                \hline 
           \multirow{2}{*}{SNI}&Instruct-T5 & 0.751& 0.751 \\
           %\hline
           &FLAN-T5 & 0.873 & 0.883 \\
           \bottomrule

    \end{tabular}
    \end{sc}
    \end{small}
    \caption{AUROC of well-calibrated models with and without post-hoc calibration at training time. We find that there is no notable performance increase with post-hoc calibration, and that there is a performance \textit{decrease} when the model has previously been tuned on the calibration data.}
    \label{tab:posthoc}
\end{table*}

\section{SuperNatural-Instructions Tasks}
\label{app:sni-data}

\subsection{Target Calibration Tasks}
\label{app:sni-data-calib}

As we describe in \autoref{sec:method}, in this work we rely on the assumption that our target-tasks have a correct answer, in the sense that it can be easily verified that an answer is right or wrong.
Although this is not a strict necessity for calibration, it allows for us to define our buckets in terms of expected accuracy, rather than e.g. an expected score. 
We therefore focus on {\it short-form} Q\&A tasks, question-answer pairs whose answers consist of either selection from a fixed answer set (e.g. multiple choice or fixed choice) or single-word answers.
We identify $15$ tasks from the SuperNatural-Instructions dataset~\citep{wang-etal-2022-super} that fit our criteria, and hold out these tasks as our uncertainty prediction tasks.

These tasks are split across $4$ rough task types: {\bf Multiple Choice} tasks involve selecting an answer from a set of choices, where the response is either a number or letter indicating the choice; {\bf Fixed Choice} tasks involve selecting an answer from a pre-defined set of choices that are constant across the task (e.g. respond with either \texttt{True} or \texttt{False}); {\bf Span Selection} tasks involve selecting the correct span of text from context and responding with that span as the answer; {\bf Open Answer} involves generating the answer to the question in an open-ended way, i.e. the answer is not provided in the context.

For all tasks, we ensure that the answers are no more than 2 words long, making it easy to perform normalization and verify accuracy for each question.
The tasks are shown in \autoref{table:calib-test-tasks}; for each task, we use $10\%$ of the samples as a validation set, $10\%$ of the samples as a held-out test set, use the remaining $80\%$ of the data to form our calibration set.

\begin{table*}[h]
    \centering
    \begin{tabular}{c|c}
    \toprule
          Task Type  & Task Name \\
    \midrule
        \multirow{9}{*}{Multiple Choice}
        & \texttt{task580-socialiqa-answer-generation} \\
        & \texttt{task309-race-answer-generation} \\
        & \texttt{task1297-qasc-question-answering} \\
        & \texttt{task1420-mathqa-general} \\
        & \texttt{task228-arc-answer-generation-easy} \\
        & \texttt{task1286-openbookqa-question-answering} \\
        & \texttt{task1431-head-qa-answer-generation} \\
        & \texttt{task1731-quartz-question-answering} \\
        & \texttt{task750-aqua-multiple-choice-answering} \\
        \midrule
        \multirow{2}{*}{Fixed Choice}
         & \texttt{task380-boolq-yes-no-question} \\
         & \texttt{task1661-super-glue-classification} \\
        \midrule
        \multirow{2}{*}{Span Selection}
        & \texttt{task002-quoref-answer-generation} \\
        & \texttt{task041-qasc-answer-generation} \\
        \midrule
        \multirow{2}{*}{Open Answer}
        & \texttt{task591-sciq-answer-generation} \\
        & \texttt{task898-freebase-qa-answer-generation} \\
        \bottomrule
    \end{tabular}
    \caption{\label{table:calib-test-tasks} The tasks and task types that we select from the SuperNatural-Instructions dataset for validating and testing our calibration method.}
\end{table*}

\subsection{Instruction-Tuning Tasks}
\label{app:sni-data-ins-tuning}

Because most modern instruction-tuned models are trained on all of Super-NaturalInstructions, they have seen the our calibration target tasks during instruction-tuning.
Therefore, we instruction-tune our own T5 model to test the effectiveness of our method on unseen tasks.
Our model is trained on a subset of the SuperNatural-Instructions dataset~\citep{wang-etal-2022-super}.
Specifically, we instruction-tune on the English split used in the original paper but we take out our target calibration tasks identified in \autoref{app:sni-data-calib}.
This gives us a training dataset of $879$ instruction-tuning tasks, with a total of roughly 1.2M training samples total.

To validate our models instruction-following capabilities, we use the in-context learning test set from SuperNatural-Instructions, which contains 95 additional held out tasks from task categories that are not seen in the training dataset.
\section{Ministral plots}\label{sec:more_plots}
\begin{figure*}[h]
\centering
\includegraphics[scale=.21]{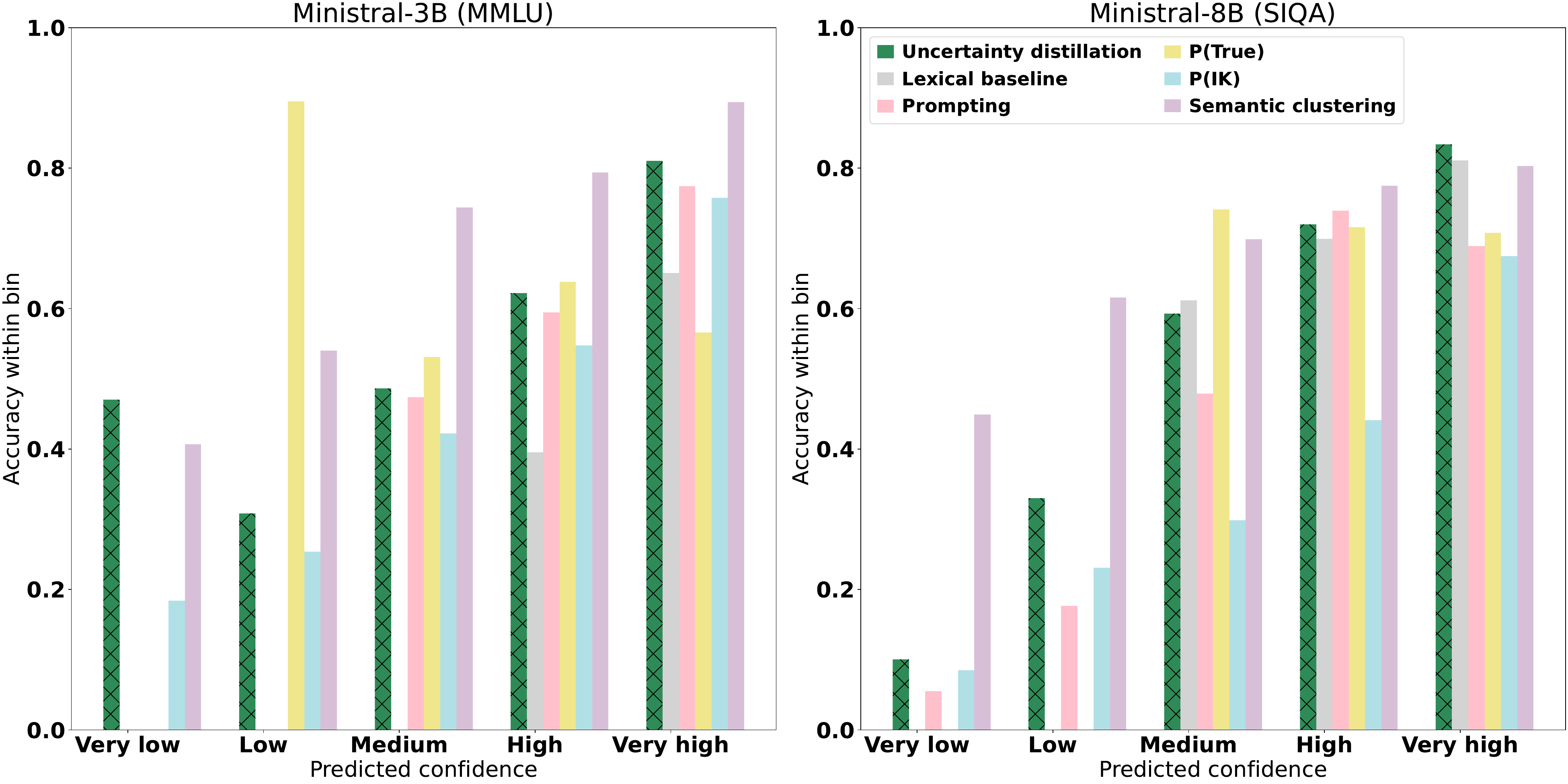}
  \caption{\textcolor{black}{Average accuracy within each confidence bin for our main experiments. We do not plot bins with fewer than 10 samples.}}\label{fig:mistral}
\end{figure*}

In \autoref{fig:mistral} we display the plots with Ministral-8B. As reflected in the AUROC score in \autoref{tab:main}, calibration is slightly worse; however, compared to baselines, it still does a more accurate job of forecasting accuracy.

\section{Instruction-Tuning T5}\label{sec:instruct-hparams}

We follow a standard recipe for instruction-tuning T5-Large, established in \citet{wang-etal-2022-super}. Specifically, we tune the model for 3 epochs with a batch size of $16$ and a learning rate of $5\times10^{-3}$. We use the AdamW optimizer, and a constant learning rate schedule after a warmup period of $500$ steps.
During instruction-tuning, we train the model with the semantic definition of each task prepended to the task input, and we similarly prompt the model when performing our target Q\&A tasks.

%\section{Help Me}
%\section{Stupid Stuff}
\section{Resource Reporting}\label{sec:resources}

\subsection{Compute Resources}

Here we report the compute resources used in this work.
Instruction-tuning T5 took a total of $200$ GPU hours across $4$ NVIDIA-V100s. Running uncertainty distillation on Instruct-T5 and FLAN-T5 took 16 hours per model on a single NVIDIA-H100. Finetuning T5-base on SQUAD for our initial model took 3 hours on a single NVIDIA RTX 2080, and training using uncertainty distillation took 8 hours on a single NVIDIA-V100. Finetuning Ministral-8B (LoRA) and finetuning Llama-3B each took took three hours on two NVIDIA-A100s. Our lexical baseline for SQUAD took one hour on one NVIDIA RTX 2080; for SNI took three hours on one NVIDIA RTX 2080; for MMLU took three hours on one NVIDIA-A100; for SocialIQA took three hours on one NVIDIA-A100; \textcolor{black}{for GSM8K took four hours on one NVIDIA-A100}. Prompting for MMLU and prompting for SocialIQA took 1 hour on one NVIDIA-A100. Sampling for SQUAD took a total of 60 GPU hours on NVIDIA-V100s; for SNI took 45 GPU hours on NVIDIA-A100s; for SocialIQA took 350 hours on NVIDIA-A100s; for MMLU took 350 hours on NVIDIA-A100s; \textcolor{black}{for GSM8k took 80 hours on NVIDIA-A100s}.

\subsection{Resource Intended Use}

Super-NaturalInstructions (SNI) is an open-source instruction tuning dataset, released under the Apache License.\footnote{Available here: \url{https://github.com/allenai/natural-instructions}}
The intended use of SNI is to instruction-tune language models to learn to follow instructions, and to evaluate a model's ability to follow instructions on unseen tasks.
While we use the SNI dataset for precisely this purpose during instruction-tuning, we also use $15$ held-out tasks to serve as uncertainty quantification tasks.
This does not necessarily fall under the intended use of instruction-tuning; however, the authors of SNI also mention that the dataset may serve as a large, multi-task natural language resource~\citep{wang-etal-2022-super}, and our usage of the target calibration tasks does fall under this use case.

The Stanford Question Answering Dataset (SQUAD)~\citep{rajpurkar2016squad} is distributed under the  Creative Commons Attribution-Sharealike 4.0 license, which permits use of the dataset as long as it is properly attributed and as long as the results are distributed under the same license. As we cite the paper and plan to publically release our code and models after acceptance, our use of this dataset is permitted under this license.

SocialIQA~\citep{siqa} is not explicitly licensed, but they state that they `` establish Social IQa as a resource'' for future models.

MMLU~\citep{mmlu}, \textcolor{black}{GSM8K\citep{gsm8k}, and MMLU-pro\citep{wang2024mmluprorobustchallengingmultitask}} are published under the MIT license, which allows users to freely copy, use, and change the licensed material.

\section{Uncertainty distillation hyperparameters}\label{sec:hparams}
In \autoref{tab:hparams} and \autoref{tab:hparams2}, we show the training hyperparameters for uncertainty distillation training. All experiments in \autoref{tab:hparams} added two incorrect answers per question, and in \autoref{tab:hparams2} added one incorrect answer per question.

\textcolor{black}{\paragraph{Hyperparameters for fine-tuning via API} For the experiments reported in~\autoref{sec:api-tuning}, we fine-tune the \texttt{gemini-2.5-flash-lite} model using LoRA with rank $4$ for $10$ epochs and defaults for other hyperparameters. On validation data, we compared performance for different numbers of incorrect examples (\autoref{sec:selfannote}), finding that augmenting the tuning set with a single incorrect prediction had marginal impact on accuracy while significantly improving calibration. We also compared both temperature scaling and isotonic regression, finding that isotonic scaling produced better calibration, while temperature scaling produced higher accuracy. To fit the calibration map, we held out 10\% of the training data.}

\begin{table*}[h]
    \centering
    \begin{tabular}{c|cccc}
    \toprule
          Model & Epochs &Learning rate & Batch size& Grad accum steps   \\
         \midrule
         T5-base (initial)&
          1&3e-5  &12&1 \\ 
          T5-base (Uncertainty distillation)&
          3&3e-5  &12&1 \\ 
          Instruct-T5 (Uncertainty distillation) &
          3&3e-5  &1&32 \\ 
          FLAN-T5 (Uncertainty distillation)&
          3&3e-5  &1&32 \\ \bottomrule

    \end{tabular}
    \caption{Hyperparameters for training all T5 models but Instruct-T5 (see \autoref{sec:instruct-hparams} for details). All models are trained with the AdamW optimizer.}
    \label{tab:hparams}
\end{table*}

\begin{table*}[h]
    \centering
    \begin{tabular}{c|ccccc}
    \toprule
          Model & Epochs &Learning rate & Batch size& LoRA rank & LoRA alpha  \\
         \midrule
         Llama-3B/MMLU&
          3&4e-5&4&-&- \\ 
          Llama-3B/SocialIQA&
          1&3e-5&4& -& -\\ 
          Ministral-8B/MMLU&
           3 & 5e-5 &4 &16&32\\ 
          Ministral-8B/SocialIQA&
          1&3e-5&4&8&16 \\ 
          \color{black}Llama-3B/GSM8K&
          \color{black}1&\color{black}8e-6&\color{black}4&\color{black}-&\color{black}- \\\bottomrule

    \end{tabular}
    \caption{Hyperparameters for training all Llama and Ministral models. Gradient accumulation steps is 1 for each model. All models are trained with the AdamW optimizer. }
    \label{tab:hparams2}
\end{table*}
\section{Algorithm}

\begin{algorithm}
\caption{Uncertainty distillation}\label{alg:method}
\begin{algorithmic}

\Require Language model $f_\theta$ with params $\theta_0$
\Require Calibration set $S^{cal} = \{ X^{cal}, Y^{cal} \}$
\State $S^{scored} \leftarrow \emptyset$
\For{$(x, y) \in S^{cal}$}
\State $D \leftarrow \{\hat{y}_i\}_{i=1}^N \sim f_\theta(x)$
\State Normalize $D$ by semantics, and count
\For{$\hat{y} \in D$ with count $n$}
\State $f \leftarrow \frac{n}{N}$
\State $S^{scored} \leftarrow S^{scored} \cup \{(x, \hat{y}, y, f) \}$
\EndFor
\EndFor
\State $c() \leftarrow $ \texttt{isotonic\_regression}$(S^{scored})$
\State $S^{vc} = \emptyset$
\For{$(x, \hat{y}, y, f) \in S^{scored}$}
\If{\texttt{filter}$(\hat{y}, y)$}
\State \texttt{continue}
\EndIf
\State $p \leftarrow c(f)$
\State $b \leftarrow$ \texttt{bin}$(p)$
\State $z \leftarrow$ \texttt{verbalize\_confidence\_map}$(\hat{y}, b)$
\State $S^{vc} \leftarrow S^{vc} \cup \{ (x, z )\}$

\EndFor
\State $ \mathcal{L}(\theta) \leftarrow \mathbb{E}_{(x, z) \in S^{vc}} [ NLL(f_\theta(x), z) ]$
\State $\theta_{cal} \leftarrow$ \texttt{train}$(\theta_0, \mathcal{L})$
\State Return $\theta_{cal}$
\end{algorithmic}
\end{algorithm}

\end{document}